\documentclass[twoside,11pt]{article}
\usepackage{jmlr2e}

\usepackage{amsmath}
\usepackage{hhline}
\usepackage{soul}
\usepackage{bm}
\usepackage{amssymb}
\usepackage{cite}
\usepackage[utf8]{inputenc} 
\usepackage[T1]{fontenc}    
\usepackage{url}            
\usepackage{booktabs}       
\usepackage{amsfonts}       
\usepackage{nicefrac}       
\usepackage{microtype}      

\def\Plus{\texttt{+}}
\def\Minus{\texttt{-}}

\usepackage{amssymb}
\usepackage{tcolorbox}
\usepackage{bbm}
\usepackage{leftidx}
\usepackage[labelformat=simple]{subcaption}
\renewcommand\thesubfigure{~(\alph{subfigure})}
\usepackage[normalem]{ulem}
\usepackage{enumitem}
\usepackage[linesnumbered,ruled]{algorithm2e}
\usepackage{multicol}
\usepackage[makeroom]{cancel}

\newcommand\ci{\perp\!\!\!\perp}

\usepackage{tikz}
\usetikzlibrary{decorations.pathreplacing,angles,quotes}
\usepackage{tkz-tab}
\usetikzlibrary{plotmarks}

\usepackage{pgfplots}

\pgfplotsset{compat=newest}
\usetikzlibrary{patterns}
\usepgfplotslibrary{fillbetween}

\usetikzlibrary{arrows, automata}
\usetikzlibrary{decorations.pathreplacing,angles,quotes}

\newtheorem{definition1}{Definition}
\newtheorem{theorem1}{Theorem}
\newtheorem{lemma1}{Lemma}

\newtheorem{assumption1}{Assumption}

\newtheorem{innercustomthm}{Theorem}
\newenvironment{customthm}[1]
  {\renewcommand\theinnercustomthm{#1}\innercustomthm}
  {\endinnercustomthm}

\makeatletter
\def\ps@pprintTitle{%
   \let\@oddhead\@empty
   \let\@evenhead\@empty
   \let\@oddfoot\@empty
   \let\@evenfoot\@oddfoot
}
\makeatother

\makeatletter
\newcommand*\rel@kern[1]{\kern#1\dimexpr\macc@kerna}
\newcommand*\widebar[1]{%
  \begingroup
  \def\mathaccent##1##2{%
    \rel@kern{0.8}%
    \overline{\rel@kern{-0.8}\macc@nucleus\rel@kern{0.2}}%
    \rel@kern{-0.2}%
  }%
  \macc@depth\@ne
  \let\math@bgroup\@empty \let\math@egroup\macc@set@skewchar
  \mathsurround\z@ \frozen@everymath{\mathgroup\macc@group\relax}%
  \macc@set@skewchar\relax
  \let\mathaccentV\macc@nested@a
  \macc@nested@a\relax111{#1}%
  \endgroup
}
\makeatother

\begin{document}

\firstpageno{1}

\title{Dirac Delta Regression:\\ Conditional Density Estimation with Clinical Trials}

\author{\name Eric V. Strobl
       \AND
       \name Shyam Visweswaran}

\editor{TBA}
\maketitle

\begin{abstract}
Personalized medicine seeks to identify the causal effect of treatment for a particular patient as opposed to a clinical population at large. Most investigators estimate such personalized treatment effects by regressing the outcome of a randomized clinical trial (RCT) on patient covariates. The realized value of the outcome may however lie far from the conditional expectation. We therefore introduce a method called \textit{Dirac Delta Regression} (DDR) that estimates the entire conditional density from RCT data in order to visualize the probabilities across all possible outcome values. DDR transforms the outcome into a set of asymptotically Dirac delta distributions and then estimates the density using non-linear regression. The algorithm can identify significant differences in patient-specific outcomes even when no population level effect exists. Moreover, DDR outperforms state-of-the-art algorithms in conditional density estimation by a large margin even in the small sample regime. An R package is available at https://github.com/ericstrobl/DDR.
\end{abstract}
\begin{keywords}
  Randomized Clinical Trial, Conditional Density Estimation, Causality, Small Sample Regime
\end{keywords}

\section{The Problem} \label{sec_intro}
\textit{Randomized clinical trials} (RCTs) are the gold standard for inferring causal effects of treatment in biomedicine. These trials utilize the notion of \textit{potential outcomes}, where we consider a set of treatments $\mathcal{T}$ and a random variable $T$ denoting the treatments assigned from $\mathcal{T}$. We then postulate the existence of a random vector $\bm{Y} \triangleq (Y(t))_{t \in \mathcal{T}}$ representing the causal outcomes of all treatments \citep{Rubin74}. The density $f(Y(t))$ intuitively summarizes the probabilities of all possible outcome values for $T=t$, so we seek to estimate $f(Y(t))$ for each treatment $t \in \mathcal{T}$. We can solve this problem by directly measuring $\bm{Y}$ in an ideal world. However, we can only assign each patient to a single treatment $T=t$ and observe $Y(T=t)$ in practice. We thus can only sample from $f(Y(T)|T)$, even though we want to sample from $f(\bm{Y})$. RCTs circumvent this problem by randomizing $T$ so that $T$ and $\bm{Y}$ are probabilistically independent, or \textit{unconfounded}. The independence implies $f(Y(t)|T) = f(Y(t))$ for any $t \in \mathcal{T}$ which in turn implies $f(Y(T)|T) = f(Y(T))$. As a result, we can use RCT data to estimate $f(Y(t))$ for each $t \in \mathcal{T}$.

Biomedical investigators have traditionally attempted to summarize $Y(T)$ by estimating the aforementioned densities or computing simple statistics of $Y(T)$, such as the mean or variance, which average over patients. However, averages fail to capture patient heterogeneity when patients within the same clinical population respond to treatment differently. For example, all patients with breast cancer do not respond better to the medication raloxifene ($T=1$) relative to placebo ($T=0$), but patients who are estrogen or progesterone receptor positive often do \citep{Martinkovich14}; the densities $f(Y(0))$ and $f(Y(1))$ therefore only differ when restricted to breast cancer patients with particular molecular receptors in this case. Biomedical investigators are thus also interested in computing statistics of $Y(T)$ \textit{given} a set of patient covariates $\bm{X}$ in order to elucidate patient-specific outcomes.

Most investigators in particular estimate the \textit{conditional expectation} by regressing $Y(T)$ on $\bm{X}$ (e.g., \citep{Zhang12,Luedtke16,Zhang2012c}). The conditional expectation however does not capture the uncertainty in outcome value because it assigns a single point prediction $\mathbb{E}(Y(T)|\bm{X}=\bm{x})$ to each patient. The realized outcome value for a particular patient may lie far from the conditional expectation. As a result, patients and healthcare providers instead prefer to know the conditional probabilities associated with \textit{all} possible values of $Y(T)$ in order to make more informed clinical decisions \citep{Akobeng07,Diamond79,Bhise18,Mah16,Berger15}.

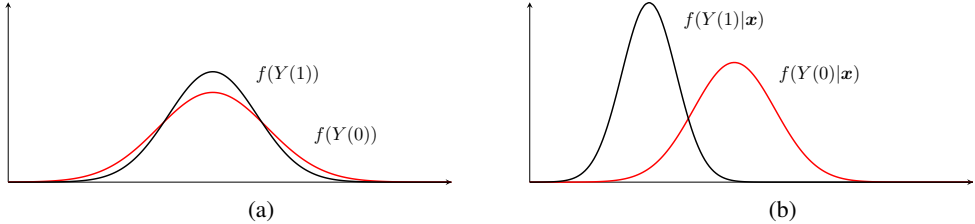
\begin{figure*}
\centering
\pgfmathdeclarefunction{gauss}{2}{%
    \pgfmathparse{1/(#2*sqrt(2*pi))*exp(-((x-#1)^2)/(2*#2^2))}%
}
\begin{subfigure}[b]{0.49\textwidth}
\centering
    \begin{tikzpicture}[scale=0.7,trim left=0.75cm]

    \def\startx{-6}    
    \def\endx{7}       
    \def\camean{0}  
    \def\casigma{1.6}  
    \def\cbmean{0}   
    \def\cbsigma{1.3}  
    \def\verticala{-6} 
    \def\verticalb{6} 

    \begin{axis}[
    domain=\startx:\endx,
    samples=101,
    ymax=0.5,
    enlargelimits=false,
    axis x line=middle,
    axis y line=left,
    xtick={\empty},
    ytick={\empty},
    height=5cm,
    width=10cm
    ]
    \addplot [name path=a,thick, smooth, red] {gauss(\camean,\casigma)};
    \addplot [name path=b,thick, smooth] {gauss(\cbmean,\cbsigma)};%
    \path[name path=axis] (axis cs:\startx,0) -- (axis cs:\endx,0);
    \node (a) at (3.9,0.13) {$f(Y(0))$};
    \node (b) at (2.2,0.3) {$f(Y(1))$};

    \end{axis}

    \end{tikzpicture}
    \caption{}\label{fig_densities:population}
    \end{subfigure}
\begin{subfigure}[b]{0.49\textwidth}
\centering
    \begin{tikzpicture}[scale=0.7,trim left=0.75cm]

    \def\startx{-6}    
    \def\endx{7}       
    \def\camean{0}  
    \def\casigma{1.2}  
    \def\cbmean{-2.5}   
    \def\cbsigma{0.8}  
    \def\verticala{-6} 
    \def\verticalb{6} 

    \begin{axis}[
    domain=\startx:\endx,
    samples=101,
    ymax=0.5,
    enlargelimits=false,
    axis x line=middle,
    axis y line=left,
    xtick={\empty},
    ytick={\empty},
    height=5cm,
    width=10cm
    ]
    \addplot [name path=a,thick, smooth, red] {gauss(\camean,\casigma)};
    \addplot [name path=b,thick, smooth] {gauss(\cbmean,\cbsigma)};%
    \path[name path=axis] (axis cs:\startx,0) -- (axis cs:\endx,0);
    \node (a) at (-0.3,0.45) {$f(Y(1)|\bm{x})$};
    \node (b) at (2.5,0.3) {$f(Y(0)|\bm{x})$};

    \end{axis}

    \end{tikzpicture}
    \caption{} \label{fig_densities:patient}
    \end{subfigure}

    \caption{Treatments $0$ and $1$ are clinically equivalent according to the  unconditional densities shown in (a). However, a patient with covariates $\bm{X}=\bm{x}$ will likely benefit more from treatment $1$ as shown by the reduction in symptom severity with the conditional densities in (b).} \label{fig_densities}
\end{figure*}

Conditional densities can fortunately summarize the desired probabilities in a single intuitive graph, especially when the potential outcomes are univariate. We therefore consider recovering $f(Y(T)|\bm{X})$ from RCT data. Consider for example two treatments $\mathcal{T} = \{0,1\}$. We can estimate the unconditional densities $f(Y(0))$ and $f(Y(1))$ using RCT data as shown in Figure \ref{fig_densities:population}. Both of the densities are very similar because they summarize the treatment outcomes across everyone recruited in the RCT, some of whom may respond well to treatment while others may not. In contrast, Figure \ref{fig_densities:patient} displays the densities $f(Y(0)|\bm{x})$ and $f(Y(1)|\bm{x})$ estimated from the original RCT, as if we had run an RCT personalized towards a particular patient where everyone recruited into the trial had the exact same covariate values $\bm{X}=\bm{x}$. Notice that most of the patients in this trial who received $T=1$ had a reduction in symptom severity, as measured using univariate outcomes, compared to those who received $T=0$. Conditional densities thus can summarize the results of a personalized RCT which may differ substantially from the results of the original RCT.

Conditional densities are unfortunately much more difficult to estimate than conditional expectations in the non-parametric setting, especially with the small sample sizes seen in RCTs (e.g., at most a few hundred samples per treatment). Existing methods struggle in this setting.
\begin{tcolorbox}
We therefore propose a new procedure called Dirac Delta Regression (DDR) that accurately estimates non-parametric conditional densities even in the small sample regime in order to intuitively visualize the probabilities associated with all possible outcome values using RCT data.
\end{tcolorbox}
\noindent DDR transforms $Y(T)$ into a set of asymptotically Dirac delta distributions and then regresses the distributions on $\bm{X}$ as described in the Section \ref{sec_alg}. The algorithm also \textit{directly} estimates $f(Y(T)|\bm{X})$ using established regression procedures and tunes a minimal number of hyperparameters automatically without access to the true conditional density. We prove consistency of the procedure as well as establish the bias rate with respect to the transformation in Section \ref{sec_theory}. Experiments show that DDR (a) leads to state-of-the-art performance in practice and (b) allows biomedical investigators to estimate the densities of a personalized RCT from the original RCT data without additional data collection. 

\section{Related Work} \label{sec_related}

We can estimate non-parametric conditional densities using several other methods besides DDR in order to visualize the probabilities associated with all values of $Y(T)$. Conditional Kernel Density Estimation (CKDE) for example estimates $f(Y(T)|\bm{X})$ using the ratio $f(Y(T),\bm{X})/f(\bm{X})$, where the numerator and denominator are approximated using kernel density functions \citep{Rosenblatt69,Hyndman96,Fosgerau10}. Other algorithms first estimate the conditional cumulative distribution function (CDF) of $Y(T)$ given $\bm{X}$ and then recover the associated density by smoothing the estimated CDF \citep{Takeuchi06,Cattaneo18}. These indirect approaches to estimating $f(Y(T)|\bm{X})$ however degrade accuracy in practice because they can incur error when estimating the joint density or the CDF. 

Investigators therefore  later proposed to directly estimate  $f(Y(T)|\bm{X})$ instead. The earliest methods in this category, such as the Mixture Density Network (MDN), estimate $f(Y(T)|\bm{X})$ by performing expectation-maximization over a mixture of parametric densities \citep{Bishop06,Tresp01}. These methods however are very time consuming, and their accuracy depends heavily on the choice of the component densities. \citet{Sugiyama10} and \citet{Kanamori12} thus proposed the Least Squares Conditional Density Estimation (LSCDE) algorithm which estimates $f(Y(T)|\bm{X})$ in closed form using reproducing kernels. The authors nevertheless found that LSCDE performs poorly when $\bm{X}$ contains multiple variables. 

Researchers suggested handling the higher dimensional setting by integrating both dimension reduction and direct density estimation into a single procedure \citep{Shiga15,Izbicki16,Tangkaratt15}. These methods, such as Series Conditional Density Estimation (SCDE), outperform LSCDE on average when sparsity or a low dimensional manifold exist, but they still struggle to accurately estimate $f(Y(T)|\bm{X})$ when the signal does not follow a simple structure. As a result, \citet{Izbicki17} introduced an algorithm called FlexCode (FC) which utilizes non-linear regression on an orthogonal series (e.g., the Fourier series). FC admits a variety of regression procedures and therefore can capitalize on the successes of regression in high dimensional estimation. Many of the proposed regression procedures however only help FC produce accurate estimates of $f(Y(T)|\bm{X})$ in special cases, so it is often unclear how to choose the best regressor in practice. FC also tends to over-smooth its conditional density estimates by enforcing a small number of orthogonal bases. The method therefore can have trouble handling the complexities inherent in real data like its predecessors.

The normalizing flow estimator (NFE) attempts to directly estimate $f(Y(T)|\bm{X})$ in the high dimensional setting using a different strategy called the normalizing flow \citep{Trippe18}. A normalizing flow automatically ensures that the density integrates to one by transforming an initial density with a series of invertible transformations using a deep Bayesian neural network. NFE must tune its prior distributions, flow parameters and network architecture in addition to the network weights. The algorithm therefore tends to overfit in practice,  especially in the small sample regime seen with the majority of RCTs.

In this paper, we improve upon the aforementioned works by proposing a method called DDR that (1) estimates $f(Y(T)|\bm{X})$ directly without approximating $f(Y(T),\bm{X})$ first, (2) handles the high dimensional setting using established regression procedures, and (3) balances overfitting and underfitting even with small RCTs using an automated cross-validation procedure over few hyperparameters. We summarize the differences between DDR and prior approaches in Table \ref{table_comp}. 

\begin{table*}
\footnotesize
\centerline{
\begin{tabular}{lcccc}
\hhline{=====}
                   & Small samples & Direct estimation & Non-parametric  & High dimensions  \\ \hline
                  
                  CKDE & & & \checkmark  & \\
                  MDN & & \checkmark & & \\
                  LSCDE & & \checkmark & \checkmark & \\
                  NFE & & \checkmark & \checkmark & \\
                  SCDE & & \checkmark & \checkmark & \checkmark \\
                  FC & & \checkmark & \checkmark & \checkmark  \\
                  DDR & \checkmark & \checkmark & \checkmark & \checkmark \\

\hhline{=====}
\end{tabular}
}
\caption{Properties of prior approaches compared to DDR.} \label{table_comp}
\end{table*}

\section{Algorithm Design} \label{sec_alg}

\subsection{Setup}
We now provide a detailed description of the DDR algorithm. We first consider independent and identically distributed (i.i.d.) data in the form of triples $\{(\bm{x}_i, t_i, y_i(t_i))\}_{i=1}^n$ collected from an RCT. The data corresponds to instantiations of the random variables $(\bm{X}, T, Y(T))$, where $Y(T)$ is assumed to be continuous, and $\bm{X}$ denotes a vector of covariates (continuous, discrete or both) measured before treatment assignment. We assume that we have $f(T)>0$. We also adopt the \textit{strongly ignorable treatment assignment} (SITA) or \textit{unconfoundedness} assumption in the conditional setting which asserts that we have $T \ci \{Y(t)\}_{t \in \mathcal{T}} | \bm{X}$ \citep{Rosenbaum83,Rubin90}. The SITA assumption holds with an RCT because investigators must assign treatments independently of $Y(T)$ within any stratum of the covariates. The assumption also implies that we have:
\begin{equation} \nonumber
\begin{aligned}
f(Y(T)|T,\bm{X})=f(Y(T)|\bm{X}),
\end{aligned}
\end{equation}
so that the latter term can be estimated from RCT data. In this paper, we seek to estimate the density $f(Y(t)|\bm{X})$ for each treatment value $t \in \mathcal{T}$ using DDR.

\subsection{Dirac Delta Regression} \label{sec_DDR}

We design DDR for the setting where $Y(t)$ is continuous. However, it is informative to first consider the discrete setting. Suppose that the discrete variable $Y(t)$ takes on $d$ distinct values in the set $\{z_j\}_{j=1}^d$. Recall that we have $\mathbb{E}[\mathbbm{1}(Y(t)= z_k)|\bm{X}]=f_{Y(t)}(z_k|\bm{X})$ for each $z_k \in \{z_j\}_{j=1}^d$. We can therefore estimate the probability mass function $f(Y(t) | \bm{X})$ by first transforming $Y(t)$ into a set of $d$ indicator functions. In particular, we compute $\mathbbm{1}(Y(t) = z_k)$ for each $z_k \in \{z_j\}_{j=1}^d$. With $n$ samples, we thus convert a column vector containing $n$ samples of $Y(t)$ into an $n \times d$ matrix with the $i,k^\textnormal{th}$ entry corresponding to $\mathbbm{1}(y_i(t)= z_k)$. We finally proceed with regressing the set of $d$ indicators on $\bm{X}$ to obtain an estimate of $\mathbb{E}[\mathbbm{1}(Y(t)= z_k)|\bm{X}]$ for each $z_k \in \{z_j\}_{j=1}^d$.

The quantity $\mathbbm{1}(Y(t) = z_k)$ is unfortunately almost surely equal to zero in the continuous case. The aforementioned strategy therefore fails in this setting because the $n \times d$ matrix is a matrix of zeros almost surely. Recall however that the following relation holds when $Y(t)$ is continuous \citep{Kanwal11}:
\begin{equation} \nonumber
    f_{Y(t)}(z_k|\bm{X}) = \mathbb{E}[\delta(z_k - Y(t)) | \bm{X}],
\end{equation}
where $\delta(z_k - Y(t))$ denotes a Dirac delta distribution, and $\{z_j\}_{j=1}^d$ is now an arbitrary set of values on $\mathbb{R}$ (we will specify a practical choice of this set in the next subsection). We can loosely view $\delta(z_k - Y(t))$ as a function equal to infinity on a set of Lebesgue measure zero:\footnote{Note that Dirac delta is more formally a generalized function or distribution, but we aim to provide the intuition here.}
\begin{equation} \nonumber
\delta(z_k - Y(t)) = \begin{cases}
+\infty & \textnormal{if }Y(t)=z_k,\\
0 & \textnormal{otherwise}.
\end{cases}
\end{equation}
As a result, we cannot transform the response variable into a set of Dirac delta distributions without again obtaining an $n \times d$ matrix of zeros almost surely. We can however ``stretch out'' the support of the Dirac delta distribution from a single point to an interval by utilizing \textit{asymptotically} Dirac delta distributions $\delta_h$ with the stretching parameter $h$ such that $\lim_{h \rightarrow 0^+} \delta_h(z_k-Y(t)) = \delta(z_k-Y(t))$. We in particular choose to utilize a Gaussian density $\delta_h(z_k-Y(t)) = \frac{1}{\sqrt{2\pi}h}\textnormal{exp}\Big(-\frac{|z_k-Y(t)|^2}{2h^2}\Big)$ for an unbounded outcome, or a truncated Gaussian for a bounded outcome, although many other densities are also appropriate; we may for example utilize the biweight or tricube density instead. As a general rule, densities with mean zero and lower higher order moments tend to perform better in practice. We can therefore recommend many kernel density functions $\frac{1}{h}K\Big(\frac{z_k-Y(t)}{h}\Big)$ used in non-parametric unconditional density estimation as well \citep{Wand94}. Regardless of our choice of $\delta_h$, we transform $Y(t)$ into a set of $d$ asymptotically Dirac delta distributions, or kernel density functions, by computing $\delta_h(z_k - Y(t))$ for each $z_k \in \{z_j\}_{j=1}^d$. We therefore convert a column vector containing the $n$ samples of $Y(t)$ into an $n \times d$ matrix with the $i,k^\textnormal{th}$ entry corresponding to $\delta_h(z_k - y_i(t))$. 

Let $\Delta_h = \{ \delta_h(z_j-Y(t))\}_{j=1}^d$ denote the set of $d$ kernel density functions. DDR then proceeds with non-linear regression of $\Delta_h$ on $\bm{X}$ using the training set in Step \ref{alg_DDR:regress_dirac} of Algorithm \ref{alg_DDR}. \textit{Thus, while DDR uses kernel density functions like CKDE, DDR directly estimates $f(Y(T)|\bm{X})$ instead of first approximating $f(Y(T),\bm{X})$ and $f(\bm{X})$ -- both of which may be much more complicated than $f(Y(T)|\bm{X})$.} DDR ultimately performs a separate regression for each $z_k \in \{z_j\}_{j=1}^d$, although it does not perform them independently due to the cross-validation procedure described in the next subsection. Informally, this strategy approximates the density values at each $z_k \in \{z_j\}_{j=1}^d$ on the test set just like with the discrete case because we have:
\begin{equation} \label{eq_intuition}
    \lim_{h \rightarrow 0^+} \lim_{n\rightarrow \infty} \widehat{g}_{h}(z_k|\bm{X}) \stackrel{p}{\rightarrow} f_{Y(t)}(z_k | \bm{X}),
\end{equation}
where $g_{h}(z_k|\bm{X}) = \mathbb{E}[\delta_h(z_k-Y(t) )| \bm{X}]$ and $\widehat{g}_{h}(z_k|\bm{X}) = \widehat{\mathbb{E}}[\delta_h(z_k-Y(t) )| \bm{X}]$ is estimated using a consistent non-linear regression method. A similar idea was proposed in \citep{Fan96}, although the authors focus on estimating the conditional expectation with local linear regression and use rules of thumb to choose $h$; we on the other hand generalize the concept to many other nonlinear regression methods and select $h$ in a principled, data dependent fashion. We formalize the intuition of Equation \eqref{eq_intuition} in Section \ref{sec_theory}, where we specifically show that the bias incurred using $\delta_h$ instead of $\delta$ is of order $h^4$ under an integrated squared error loss.

\begin{algorithm}[]
 \SetKwInOut{Input}{Input}
 \Input{\hspace{0.5mm} training set $\{ \bm{x}_i,y_i(t)\}_{i=1}^n$, test set $\{\bm{x}_i\}_{i=n+1}^m$}
 \KwResult{\hspace{0.5mm} conditional density estimates $\mathcal{G}_h \triangleq \{\widehat{g}_{h}(Y(t)|\bm{x}_i)\}_{i=n+1}^m$}
 \BlankLine

Regress $\Delta_h$ on $\bm{X}$; choose the optimal value of $h$ and all other hyperparameters $\bm{\lambda}$ using cross-validation with Equation \eqref{eq_DDR:loss_empiric} \label{alg_DDR:regress_dirac}\\

Enforce non-negativity and normalize each element in $\mathcal{G}_h$ using Equation \eqref{eq_DDR:normalize} \label{alg_DDR:neg}\\

Sharpen each element in $\mathcal{G}_{h}$ with Equations \eqref{eq_DDR:sharpen} and \eqref{eq_DDR:normalize}; choose the optimal value of $\eta$ again with Equation \eqref{eq_DDR:loss_empiric}
\label{alg_DDR:sharpen}

 \caption{Dirac Delta Regression (DDR)} \label{alg_DDR}
\end{algorithm}

\subsection{Loss Function \& Cross-Validation} \label{sec_cv}

We need to choose $h$ carefully because the stretching parameter introduces bias. If we make $h$ too large, then DDR will recover a density that is too flat. On the other hand, if we make the problem too hard with a small $h$, then the regression procedure will struggle to recover a density that is ultimately ``too wiggly.'' We therefore must pick $h$ as well as the standard set of hyperparameters of the non-linear regression method $\bm{\lambda}$ in a principled manner.

We will choose $h$ and $\bm{\lambda}$ using cross-validation. Recall however that we do not have access to the ground truth conditional density $f(Y(t)|\bm{X})$. Fortunately, we can utilize a trick by cross-validating over the Mean Integrated Squared Error (MISE) loss function \citep{Fryer76}:
\begin{align}
    &\int \int \Big| \widehat{g}_{h}(z|\bm{x}) - f_{Y(t)}(z|\bm{x}) \Big|^2~d \mathbb{P}_{\bm{X}}(\bm{x})dz \label{eq_loss1}\\
    = &\int \int
    \widehat{g}^2_{h}(z|\bm{x})~d\mathbb{P}_{\bm{X}}(\bm{x})dz \nonumber - 2 \int \int \widehat{g}_{h}(z|\bm{x}) f_{Y(t)}(z, \bm{x}) ~d\bm{x} dz + C, \nonumber
\end{align}
where $C$ is a constant that only depends on $f(Y(t)|\bm{X})$. The MISE loss thus quantifies the distance of the approximated conditional density to the true conditional density similar to the mean squared error loss for regression. The empirical MISE loss takes on the following form up to a constant:
\begin{equation} \label{eq_DDR:loss_empiric}
\begin{aligned}
    &\frac{1}{n}\sum_{i=1}^n \int \widehat{g}^2_{h}(z|\bm{x}_i)~dz - \frac{2}{n^2} \sum_{i=1}^n \widehat{g}_{h}(y_i(t)|\bm{x}_i).
\end{aligned}
\end{equation}
Notice that computing the above quantity does not require knowledge of the ground truth $f(Y(t)|\bm{X})$. We therefore can use Equation \eqref{eq_DDR:loss_empiric} to tune both $h$ and $\bm{\lambda}$ in Step \ref{alg_DDR:regress_dirac} of DDR. 

Observe however that we need to estimate $f_{Y(t)}(z|\bm{X})$ over all values of $z$ rather than just on the grid $\{z_j\}_{j=1}^d$ to compute Equation \eqref{eq_DDR:loss_empiric}. We cannot compute $f_{Y(t)}(z|\bm{X})$ over an infinite number of values, so we instead accomplish this task by linearly interpolating between the estimates $\{\widehat{g}_{h}(z_j|\bm{X})\}_{j=1}^d$. Recall that the area under the curve (AUC) of such an interpolated density converges to the AUC of the true density on any finite interval at rate $O(1/d^2)$ by the trapezoidal rule \citep{Cruz03}. We therefore set $d$ to a large number (e.g., $500$) and the grid $\{z_j\}_{j=1}^d$ to $d$ equispaced points between the maximum and minimum values of $\{y_i(t)\}_{i=1}^n$ to ensure that the interpolation error is negligible.

\subsection{Further Refinements} \label{sec_sharpen}

DDR improves upon the estimate $\widehat{g}_{h}(Y(t)|\bm{X})$ in Step \ref{alg_DDR:neg} by enforcing known properties of a conditional density such as non-negativity and a total AUC of 1; these properties do not automatically hold in the finite sample setting as in unconditional density estimation with kernel density functions because a non-linear regressor may predict negative values. The improved estimate of $f(Y(t)|\bm{X})$ thus becomes:
\begin{equation} \label{eq_DDR:normalize}
   \widehat{g}_{h}(Y(t)|\bm{X}) \leftarrow \frac{\textnormal{max}\{0, \widehat{g}_{h}(Y(t)|\bm{X})\}}{ \mathcal{T}(\widehat{g}_{h}(Y(t)|\bm{X}))},
\end{equation}
where the denominator denotes the total AUC under $\textnormal{max}\{0, \widehat{g}_{h}(Y(t)|\bm{X})\}$ as estimated using the trapezoidal rule.

Step \ref{alg_DDR:sharpen} of the DDR algorithm further modifies $\widehat{g}_{h}(Y(t)|\bm{X})$ using a parameter $\eta$ by setting:
\begin{equation} \label{eq_DDR:sharpen}
    \widehat{g}_{h}(Y(t)|\bm{X}) \leftarrow \textnormal{max}\{0, \widehat{g}_{h}(Y(t)|\bm{X})-\eta\}.
\end{equation}
We choose the optimal $\eta$ value again using Equation \eqref{eq_DDR:loss_empiric}. The $\eta$ value thus serves to eliminate low density regions of $\widehat{g}_{h}(Y(t)|\bm{X})$. Re-normalizing $\widehat{g}_{h}(Y(t)|\bm{X})$ using Equation \eqref{eq_DDR:normalize} then elongates the high density regions. As a result, Equations \eqref{eq_DDR:sharpen} and  \eqref{eq_DDR:normalize} together ``sharpen'' $\widehat{g}_{h}(Y(t)|\bm{X})$. In practice, sharpening $\widehat{g}_{h}(Y(t)|\bm{X})$ substantially improves performance in the high dimensional setting because it refines the initial conditional density estimates obtained from non-linear regression.

\begin{tcolorbox}
In summary, DDR estimates $f(Y(t)|\bm{X})$ by performing non-linear regression on a set of kernel density functions using an established regression procedure. The algorithm then refines the density estimate using Equations \eqref{eq_DDR:normalize} and \eqref{eq_DDR:sharpen}. DDR automatically tunes the associated hyperparameters $h, \bm{\lambda}$ and $\eta$ with Equation \eqref{eq_DDR:loss_empiric}. All of these improvements allow DDR to perform well even with small sample sizes and multiple dimensions.
\end{tcolorbox}

\section{Theory} \label{sec_theory}

We now establish the consistency of DDR as well as the bias of the density estimate as a function of the parameter $h$. We in particular show that the MISE loss in Equation \eqref{eq_loss1} converges to zero in probability with bias rate $O(h^4)$. 

The MISE loss is bounded above by two terms:
\begin{equation} \nonumber
\begin{aligned}
&2\int \int \Big|\widehat{g}_{h}(z|\bm{x}) - g_{h} (z|\bm{x}) \Big|^2 ~ d \mathbb{P}_{\bm{X}}(\bm{x})dz + 2\int \int \Big|g_{h} (z|\bm{x}) - f_{Y(t)}(z |\bm{x}) \Big|^2 ~ d \mathbb{P}_{\bm{X}}(\bm{x})dz
\end{aligned}
\end{equation}
The first term  corresponds to the \textit{variance} while the second corresponds to the squared \textit{bias}. We require the following assumption for the variance:
\begin{assumption1} \label{assump_reg}
The regression procedure is consistent for any fixed $h>0$ and any fixed $z$:
\begin{equation} \label{eq_part1}
Q_n(z) \triangleq \int \Big|\widehat{g}_{h}(z|\bm{x}) - g_{h} (z|\bm{x})\Big|^2 ~ d \mathbb{P}_{\bm{X}}(\bm{x}) = o_p(1).
\end{equation}
\end{assumption1}
\noindent Note that $Q_n(z)$ is a function of $\widehat{g}_{h}(z|\bm{x})$ (a stochastic regression estimate that is a function of the random dataset), and therefore $Q_n(z)$ is itself stochastic. Many non-linear regression methods satisfy the above assumption from standard results in the literature. Kernel ridge regression (KRR) for example satisfies it under some mild conditions (Corollary 5 in \citep{Smale07}). K-nearest neighbor, local polynomial, Rodeo and spectral regression also meet the requirement under their respective assumptions \citep{Izbicki17}. Assumption \ref{assump_reg} therefore allows the user to incorporate different regression procedures into DDR.

We next assume that the partial derivative of $Q_n$ with respect to (w.r.t.) $z$ is stochastically bounded:
\begin{assumption1} \label{assump_deriv}
$Q_n(z)$ is differentiable w.r.t. $z$ and $\sup_z \Big| \frac{\partial Q_n(z)}{\partial z} \Big| = O_p(1)$.
\end{assumption1}
\noindent The differentiability of $Q_n(z)$ generally holds when we choose an asymptotically Dirac delta distribution that is everywhere differentiable such as with the aforementioned Gaussian, biweight or tricube densities. The stochastic bound is equivalent to requiring a finite supremum in the deterministic setting. The assumption therefore ensures a controlled derivative across all values of $z$.

Now consider the squared bias term:
\begin{equation} \label{eq_part2}
\int \int \Big|g_{h} (z|\bm{x}) - f_{Y(t)}(z|\bm{x}) \Big|^2 ~ d \mathbb{P}_{\bm{X}}(\bm{x})dz.
\end{equation}
\noindent The above quantity depends on the choice of the asymptotically Dirac delta distribution as well as the choice of $h$. We therefore impose some additional assumptions on the distribution:
\begin{assumption1} \label{assump_kernel}
The asymptotically Dirac delta distribution is a kernel density function that admits the form $\delta_{h}(z-Y(t)) = \frac{1}{h} K \Big(\frac{z - Y(t)}{h} \Big)$ such that $\int u K(u) ~du= 0$ and $\int u^2 K(u) ~du < \infty$.
\end{assumption1}
\noindent The reader may recall an identical form used in unconditional kernel density estimation, where we also require a centered and finite variance kernel density function \citep{Vaart98}. We finally impose a mild smoothness assumption on the density $f(Y(t) | \bm{X})$ which is similarly required in the unconditional case:
\begin{assumption1} \label{assump_density}
The conditional density $f_{Y(t)}(z | \bm{X})$ is twice continuously differentiable for any $z$ and satisfies $ \int \int |f_{Y(t)}''(z |\bm{x})|^2 d\mathbb{P}_{\bm{X}}(\bm{x})dz< \infty$.
\end{assumption1}

We are now ready to state the main result:
\begin{theorem1} \label{thm_consistency}
Under Assumptions \ref{assump_reg}-\ref{assump_density}, we have:
\begin{equation} \nonumber
\int_a^b \int \Big|\widehat{g}_{h}(z|\bm{x}) - f_{Y(t)}(z |\bm{x})\Big|^2 ~ d \mathbb{P}_{\bm{X}}(\bm{x})dz \leq o_p(1) + Ch^4,
\end{equation}
for any $a<b$ where $C$ is a constant that does not depend on $n$ or $h$.
\end{theorem1}

\noindent The proof is located in Appendix \ref{sec_proofs}. Notice that the above bound holds over all possible finite intervals $[a,b]$ on $\mathbb{R}$. The parameter $h$ also imposes bias at rate $h^4$. We thus achieve consistency as $h \rightarrow 0^+$ as expected from the intuition summarized in Equation \eqref{eq_intuition}.

\section{Experiments} \label{sec_exp}

\subsection{Algorithms \& Hyperparameters}
We next compared DDR against six other algorithms:
\begin{enumerate}[leftmargin=*]
    \item LSCDE is a least squares procedure for directly estimating the conditional density \citep{Sugiyama10}. This method was shown to outperform CKDE and MDN, but LSCDE in general only performs well with a few variables in $\bm{X}$.
    \item Series Conditional Density Estimation (SCDE) improves upon LSCDE in the high dimensional setting by integrating direct conditional density estimation with dimensionality reduction \citep{Izbicki16}.
    \item FC estimates conditional densities by utilizing non-linear regression on an orthogonal series  \citep{Izbicki17}. FC admits a variety of regression procedures, so we equipped the algorithm with popular choices including Gaussian kernel ridge regression (FC-KRR), k-nearest neighbor (FC-NN) and the Lasso (FC-Lasso).
    \item Normalizing Flow Estimator (NFE) combines a deep neural network with a normalizing flow, whereby a simple initial density is transformed in a more complex one by applying a sequence of invertible transformations \citep{Trippe18}.
\end{enumerate}
Recall that we can instantiate DDR with a variety of non-linear regressors like FC. Choosing the right regressor is however a non-trivial task. To prevent the user from cherry picking and optimize performance, we strategically instantiate DDR with KRR equipped with the first degree INK spline kernel, a universal regressor that works very well in the sample size regime of nearly all RCTs (tens to few thousands) \citep{Vapnik13,Izmailov13}. The INK spline kernel is much less well known than the Gaussian one, but INK spline is tuning free and therefore does not introduce an extra hyperparameter that can increase the probability of overfitting. DDR runs in $O(dn^3)$ time in this case, and KRR allows us to compute leave-one-out predictions in closed form via the Sherman-Morrison formula to speed up cross-validation \citep{Seber12}. We selected the $h$ and $\eta$ parameters from 20 and 50 equispaced points between 0 and 0.5, respectively. We also selected the ridge penalty from $\{1\textnormal{E-}1, 1\textnormal{E-}2, 1\textnormal{E-}3, 1\textnormal{E-}4, 1\textnormal{E-}5, 1\textnormal{E-}6\}$. 

We tuned the hyperparameters of the other algorithms according to the original authors' source codes with one exception. LSCDE uses the Kullback-Leibler divergence to optimize hyperparameters by default, but we found that this method consistently resulted in worse performance using our evaluation criteria. We therefore ensured that all methods optimized their hyperparameters according to the MISE loss in order to ensure a fair comparison.

\subsection{Accuracy} \label{sec_acc}
 
 \textit{Synthetic Data.} We first evaluated the seven algorithms by simulating experimental datasets with different conditions. Most RCT datasets contain up to a few hundred samples, so we generated 200 samples of $Y(t)$ from the following models: (1) homoskedastic $\mathcal{N}(\bm{X}_1,0.1)$,(2) heteroskedastic $\mathcal{N}(\bm{X}_1, 0.1(|\bm{X}_2|+0.5))$, (3) bimodal $\frac{1}{2}\mathcal{N}(\bm{X}_1,0.1)$ + $\frac{1}{2}\mathcal{N}(\bm{X}_2,0.1)$, and (4) skewed $\bm{X}_1 + \Gamma(k=2,\theta=0.4)$. We instantiated each of the 4 models 100 times. We also set the cardinality of $\bm{X}$ to an element of $\{2,4,6,8,10,15,20\}$ by sampling i.i.d. from a standard Gaussian. We therefore compared the algorithms across a total of $4 \times 100 \times 7 = 2800$ datasets. 

\begin{figure*}
    \centering
    \begin{subfigure}[b]{0.49\textwidth}
    \centering
     \includegraphics[scale=0.51]{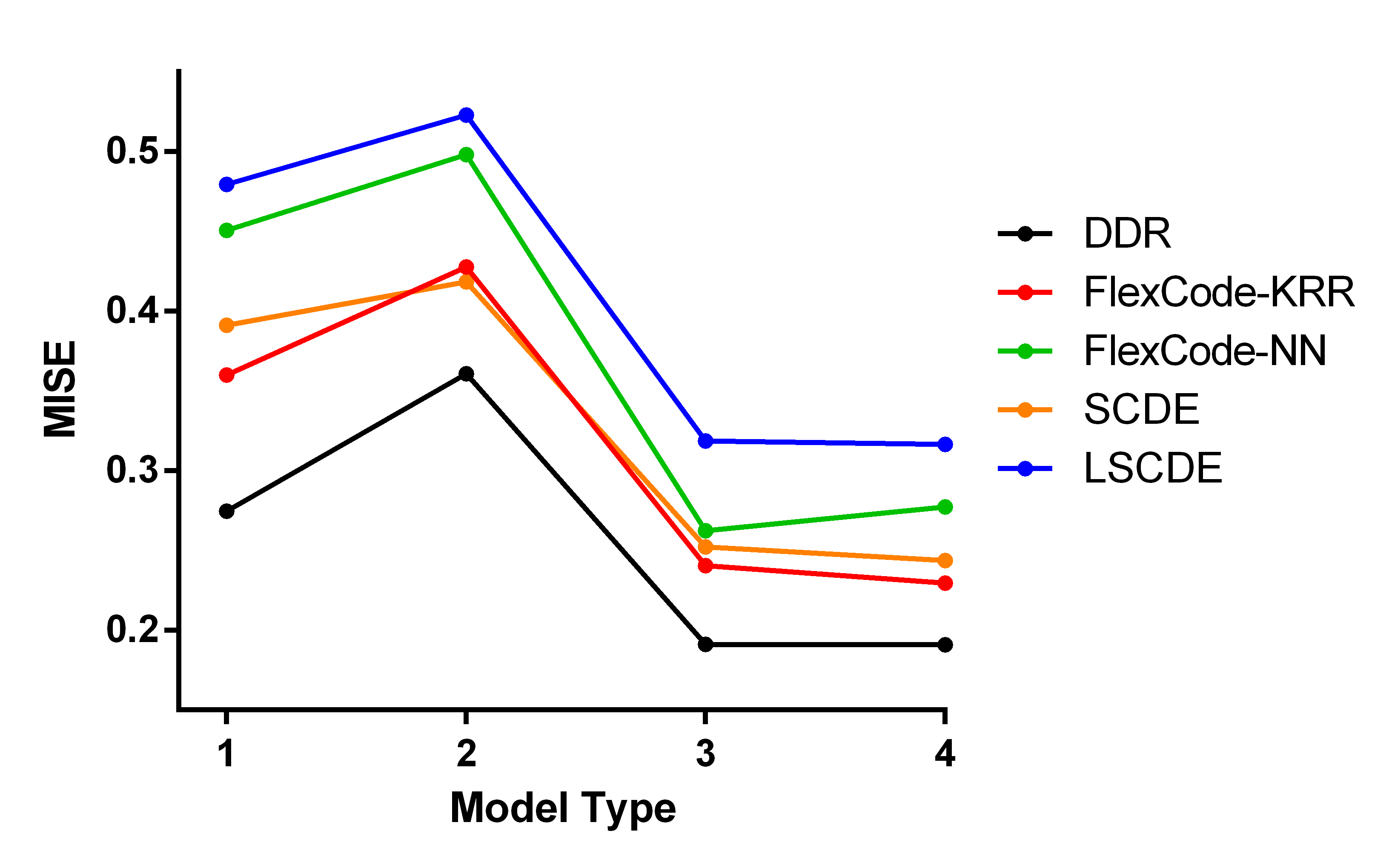}
    \caption{}
    \label{fig_synth:type}
    \end{subfigure}
     \begin{subfigure}[b]{0.49\textwidth}
        \centering
     \includegraphics[scale=0.51]{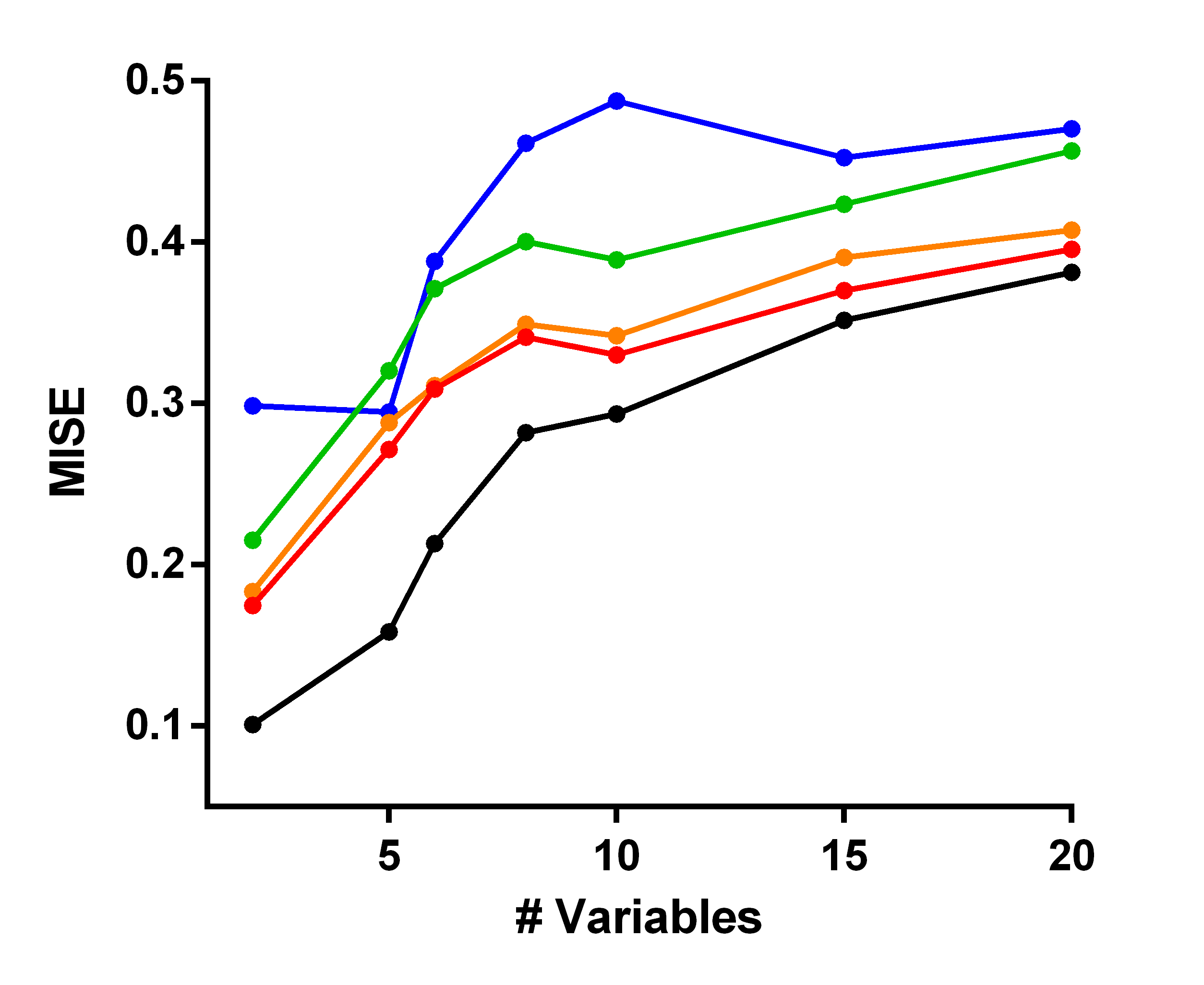}
    \caption{}
    \label{fig_synth:vars}
    \end{subfigure}
    \caption{The average MISE loss values as a function of (a) model type and (b) number of variables. We have not included FlexCode-Lasso or NFE in the plots, since they performed substantially worse. DDR achieved the lowest average MISE in all cases.} \label{fig_synth}
\end{figure*}

We evaluated the algorithms with the MISE loss in Equation \eqref{eq_loss1} as averaged over the datasets. We summarize the results in Figure \ref{fig_synth}. DDR achieved the lowest average MISE scores across all four model types (Figure \ref{fig_synth:type}) and all variable numbers (Figure \ref{fig_synth:vars}). Moreover, DDR outperformed FC-KRR to a significant degree (t=-9.57, p<2.2E-16). Every other pairwise comparison was also significant at a Bonferonni corrected threshold of 0.05/6 according to paired t-tests. FC-Lasso performed poorly, since the linear Lasso is generally not a consistent estimator of the conditional density. NFE performed the worst because deep learning usually requires large sample sizes inaccessible to most RCTs. We conclude that DDR achieves the best performance on average across a variety of situations.

 \textit{Real Data.} We next ran the algorithm on 40 real observational datasets downloaded from the UCI Machine Learning Repository with sample sizes ranging from 80 to 666. We downloaded as many pre-processed UCI datasets we could find with tens to hundreds of samples in tabular format. Predicting from observational data is equivalent to having one treatment value $T=0$ and predicting $f(Y(0) | \bm{X})$. Note that we can only compute the MISE loss up to a constant with the real data because we do not have access to the ground truths. We therefore computed Equation \eqref{eq_DDR:loss_empiric} with 10-fold cross-validation and then averaged over the folds for each dataset. Since we cannot compare the loss values between datasets, we ranked the values from 1 to 7 instead. We summarize the results in Table \ref{table_real}. DDR achieved the lowest average rank across all datasets as shown in the last row. 
 
 We repeated the above procedure with 7 clinical trial datasets downloaded from the National Institute of Drug and Alcohol (NIDA) Data Share. We computed the MISE rank for each value of $T$. DDR again achieved the lowest average rank as seen in Table \ref{table_CT}. We conclude that the results seen with both the real observational and clinical trial datasets replicate the results seen with the synthetic data.
 
 \begin{table*}[p]
\footnotesize
\centerline{
\begin{tabular}{lcccccccc}
\hhline{=========}
                                & (rank)    & \multicolumn{1}{c}{DDR}           & \multicolumn{1}{c}{FC-KRR}        & \multicolumn{1}{c}{FC-NN}         & \multicolumn{1}{c}{FC-Lasso}      & \multicolumn{1}{c}{SCDE}          & \multicolumn{1}{c}{LSCDE}         & \multicolumn{1}{c}{NFE}           \\ \hline
Autism            & 5         & \Minus1.41E\Minus01                         & \Minus2.17E\Minus01                         & \Minus6.30E\Minus01                         & \Minus6.13E\Minus01                         & \underline{\Minus6.38E\Minus01}                         & \Minus4.75E\Minus02                         & \Plus8.99E\Plus00                          \\
AutoMPG           & 4         & \Minus2.47E\Minus02                         & \Minus1.85E\Minus02                         & \Minus3.18E\Minus02                         & \Plus2.10E\Minus01                          & \underline{\Minus1.06E\Minus01}                         & \Minus2.74E\Minus02                         & \Plus2.23E\Plus01                          \\
Breast            & 4         & \Minus9.05E\Minus01                         & \Minus1.19E\Plus00                         & \Minus4.03E\Plus00                         & \Plus6.37E\Plus00                          & \underline{\Minus4.59E\Plus00}                         & \Minus9.80E\Minus02                         & \Plus1.89E\Plus00                          \\
BuddyMove         & 1         & \underline{\Minus2.23E\Minus01}                         & \Minus2.20E\Minus01                         & \Minus2.04E\Minus01                         & \Minus1.30E\Minus01                         & \Minus1.98E\Minus01                         & \Minus1.86E\Minus01                         & \Plus3.09E\Minus01                          \\
Caesarian         & 2         & \Minus9.13E\Minus03                         & \Minus7.71E\Minus03                         & \underline{\Minus1.45E\Minus02}                         & \Plus2.10E\Minus02                          & \Plus9.83E\Minus03                          & \Minus7.85E\Minus03                         & \Plus5.51E\Plus01                          \\
CKD               & 4         & \Minus3.34E\Plus00                         & \Minus5.78E\Plus00                         & \underline{\Minus7.26E\Plus00}                         & \Minus7.00E\Plus00                         & \Minus1.11E\Plus00                         & \Minus2.27E\Plus00                         & \Minus1.40E\Plus00                         \\
Coimbra           & 1         & \underline{\Minus7.15E\Minus01}                         & \Minus6.16E\Minus01                         & \Minus5.96E\Minus01                         & \Plus4.01E\Minus01                          & \Minus7.08E\Minus01                         & \Minus5.11E\Minus01                         & \Minus1.60E\Minus01                         \\
Concrete          & 1         & \underline{\Minus3.87E\Minus02}                         & \Minus3.36E\Minus02                         & \Minus2.56E\Minus02                         & \Minus1.83E\Minus02                         & \Plus6.43E\Minus02                          & \Minus3.53E\Minus02                         & \Plus1.06E\Plus02                          \\
Credit            & 2         & \Minus8.27E\Minus04                         & \underline{\Minus8.63E\Minus04}                         & \Minus7.41E\Minus04                         & \Minus2.18E\Minus04                         & \Minus4.14E\Minus04                         & \Minus6.77E\Minus04                         & \Plus8.57E\Plus02                          \\
Cryotherapy       & 3         & \Minus5.42E\Minus05                         & \underline{\Minus7.93E\Minus05}                         & \Minus6.04E\Minus05                         & \Minus8.30E\Minus06                         & \Minus3.18E\Minus06                         & \Minus3.82E\Minus05                         & \Plus3.02E\Plus05                          \\
CSM               & 1         & \underline{\Minus1.20E\Plus00}                         & \Minus1.10E\Plus00                         & \Minus1.06E\Plus00                         & \Minus7.57E\Minus01                         & \Minus1.06E\Plus00                         & \Minus1.12E\Plus00                         & \Plus8.37E\Minus02                          \\
Dermatology       & 1         & \underline{\Minus1.85E\Minus02}                         & \Minus1.49E\Minus02                         & \Minus1.20E\Minus02                         & \Minus4.52E\Minus03                         & \Minus1.33E\Minus02                         & \Minus1.38E\Minus02                         & \Plus1.11E\Plus03                          \\
Echo              & 1         & \underline{\Minus2.37E\Minus02}                         & \Minus2.35E\Minus02                         & \Minus2.20E\Minus02                         & \Minus9.02E\Minus03                         & \Minus1.19E\Minus02                         & \Minus2.32E\Minus02                         & \Plus1.72E\Plus01                          \\
EColi             & 2         & \Minus4.46E\Plus00                         & \Minus4.07E\Plus00                         & \Minus3.32E\Plus00                         & \Minus1.65E\Plus00                         & \Minus3.24E\Plus00                         & \Minus2.10E\Plus00                         & \underline{\Minus2.16E\Plus01}                         \\
Facebook          & 3         & \Minus7.08E\Minus03                         & \Minus6.31E\Minus03                         & \Minus2.94E\Minus03                         & \Minus7.94E\Minus03                         & \underline{\Minus1.87E\Minus02}                         & \Minus1.99E\Minus03                         & \Plus2.86E\Plus03                          \\
Fertility         & 1         & \underline{\Minus1.09E\Minus07}                         & \Minus8.35E\Minus08                         & \Minus7.02E\Minus08                         & \Minus1.43E\Minus08                         & \Minus1.51E\Minus08                         & \Minus6.40E\Minus08                         & \Plus1.96E\Plus07                          \\
ForestFires       & 4         & \Minus1.42E\Minus02                         & \Minus2.21E\Minus02                         & \underline{\Minus2.63E\Minus02}                         & \Plus3.18E\Minus03                          & \Plus5.13E\Minus03                          & \Minus1.69E\Minus02                         & \Plus8.57E\Plus01                          \\
ForestTypes       & 1         & \underline{\Minus2.40E\Plus01}                         & \Minus2.04E\Plus01                         & \Minus1.75E\Plus01                         & \Minus8.64E\Plus00                         & \Minus1.43E\Plus01                         & \Minus1.75E\Plus01                         & \Minus1.36E\Plus00                         \\
Glass             & 5         & \Minus6.60E\Minus02                         & \Minus7.25E\Minus01                         & \underline{\Minus8.92E\Minus01}                         & \Minus2.40E\Minus01                         & \Minus8.00E\Minus01                         & \Minus3.63E\Minus02                         & \Plus1.90E\Plus00                          \\
GPS               & 2         & \Minus9.93E\Minus03                         & \Minus8.38E\Minus03                         & \Minus8.54E\Minus03                         & \Minus9.31E\Minus03                         & \underline{\Minus1.00E\Minus02}                         & \Minus8.75E\Minus03                         & \Plus3.23E\Plus01                          \\
Hayes             & 4         & \Minus3.14E\Minus03                         & \Minus3.27E\Minus03                         & \Minus3.57E\Minus03                         & \underline{\Minus3.67E\Minus03}                         & \Minus2.88E\Minus03                         & \Minus2.53E\Minus03                         & \Plus2.43E\Plus01                          \\
HCC               & 2         & \Minus5.51E\Minus01                         & \Minus5.04E\Minus01                         & \underline{\Minus5.54E\Minus01}                         & \Minus1.60E\Minus01                         & \Plus1.67E\Minus01                          & \Minus4.07E\Minus01                         & \Plus1.65E\Plus00                          \\
Hepatitis         & 2         & \Minus6.06E\Minus02                         & \Minus4.12E\Minus02                         & \Minus3.73E\Minus02                         & \Plus1.61E\Minus03                          & \Minus2.83E\Minus02                         & \underline{\Minus8.52E\Minus02}                         & \Plus4.99E\Plus01                          \\
Immunotherapy     & 2         & \Minus1.89E\Minus01                         & \Minus1.77E\Minus01                         & \Minus1.64E\Minus01                         & \Minus1.11E\Minus01                         & \underline{\Minus2.37E\Minus01}                         & \Minus1.53E\Minus01                         & \Plus4.83E\Minus01                          \\
Inflammation      & 2         & \Minus1.45E\Minus01                         & \Minus1.08E\Minus01                         & \Minus9.99E\Minus02                         & \Minus2.48E\Minus02                         & \underline{\Minus1.86E\Minus01}                         & \Minus9.66E\Minus02                         & \Plus8.33E\Plus00                          \\
Iris              & 1         & \underline{\Minus2.26E\Plus02}                         & \Minus1.83E\Plus02                         & \Minus1.57E\Plus02                         & \Minus2.67E\Plus01                         & \Minus1.48E\Plus02                         & \Minus1.95E\Plus02                         & \Minus2.39E\Plus00                         \\
Istanbul          & 1         & \underline{\Minus1.70E\Minus02}                         & \Minus1.55E\Minus02                         & \Minus1.56E\Minus02                         & \Minus1.00E\Minus02                         & \Minus6.05E\Minus03                         & \Minus1.48E\Minus02                         & \Plus4.06E\Plus00                          \\
Leaf              & 1         & \underline{\Minus1.99E\Minus02}                         & \Minus1.90E\Minus02                         & \Minus1.24E\Minus02                         & \Minus4.70E\Minus03                         & \Minus3.53E\Minus03                         & \Minus1.13E\Minus02                         & \Plus1.36E\Plus02                          \\
Parkinson's       & 1         & \underline{\Minus6.60E\Minus02}                         & \Minus5.80E\Minus02                         & \Minus5.11E\Minus02                         & \Plus8.88E\Minus02                          & \Plus7.88E\Minus02                          & \Minus5.22E\Minus02                         & \Plus3.29E\Plus01                          \\
Planning          & 1         & \underline{\Minus1.20E\Minus02}                         & \Minus1.05E\Minus02                         & \Minus1.09E\Minus02                         & \Minus9.80E\Minus03                         & \Minus7.10E\Minus03                         & \Minus1.04E\Minus02                         & \Plus7.45E\Plus01                          \\
Seeds             & 3         & \Minus6.32E\Minus02                         & \Minus7.28E\Minus02                         & \underline{\Minus1.02E\Minus01}                         & \Plus3.38E\Minus02                          & \Plus5.63E\Minus02                          & \Minus5.30E\Minus02                         & \Plus9.99E\Plus00                          \\
Somerville        & 1         & \underline{\Minus1.48E\Minus01}                         & \Minus1.27E\Minus01                         & \Minus9.58E\Minus02                         & \Plus1.39E\Minus02                          & \Minus9.68E\Minus02                         & \Minus6.33E\Minus02                         & \Plus2.29E\Plus02                          \\
SPECTF            & 4         & \Minus3.71E\Minus01                         & \Minus4.89E\Minus01                         & \Minus5.14E\Minus01                         & \Minus3.66E\Minus02                         & \underline{\Minus6.98E\Minus01}                         & \Minus3.01E\Minus01                         & \Plus1.14E\Plus00                          \\
Statlog           & 1         & \underline{\Minus3.50E\Minus02}                         & \Minus3.11E\Minus02                         & \Minus2.41E\Minus02                         & \Minus2.54E\Minus02                         & \Plus2.05E\Minus01                          & \Minus2.19E\Minus02                         & \Plus2.25E\Plus01                          \\
Thoracic          & 1         & \underline{\Minus2.28E\Minus02}                         & \Minus2.24E\Minus02                         & \Minus2.14E\Minus02                         & \Minus2.13E\Minus02                         & \Plus4.03E\Minus02                          & \Minus1.89E\Minus02                         & \Plus5.47E\Plus01                          \\
Traffic           & 2         & \Minus1.27E\Minus01                         & \Minus1.18E\Minus01                         & \Minus9.61E\Minus02                         & \Minus5.74E\Minus02                         & \underline{\Minus1.38E\Minus01}                         & \Minus6.45E\Minus02                         & \Plus3.22E\Plus01                          \\
User              & 2         & \Minus5.45E\Minus01                         & \Minus5.21E\Minus01                         & \Minus3.47E\Minus01                         & \Minus2.90E\Minus01                         & \Minus5.03E\Minus01                         & \underline{\Minus9.38E\Minus01}                         & \Plus3.01E\Plus00                          \\
Wine              & 1         & \underline{\Minus4.70E\Minus03}                         & \Minus4.69E\Minus03                         & \Minus4.65E\Minus03                         & \Minus2.54E\Minus03                         & \Minus2.76E\Minus04                         & \Minus3.22E\Minus03                         & \Plus2.60E\Plus02                          \\
Wisconsin         & 5         & \Minus2.80E\Plus00                         & \underline{\Minus7.95E\Plus00}                         & \Minus7.03E\Plus00                         & \Minus6.54E\Plus00                         & \Minus5.93E\Plus00                         & \Minus7.63E\Minus01                         & \Plus1.68E\Plus00                          \\
Yacht             & 1         & \underline{\Minus5.12E\Plus00}                         & \Minus5.08E\Plus00                         & \Minus3.94E\Plus00                         & \Minus2.18E\Plus00                         & \Plus4.16E\Plus00  & \Minus3.12E\Plus00                         & \Minus1.46E\Minus02               \\ \hline
\textbf{Avg Rank} & & \multicolumn{1}{c}{\textbf{2.15}} & \multicolumn{1}{c}{\textbf{2.70}} & \multicolumn{1}{c}{\textbf{3.05}} & \multicolumn{1}{c}{\textbf{5.10}} & \multicolumn{1}{c}{\textbf{4.15}} & \multicolumn{1}{c}{\textbf{4.10}} & \multicolumn{1}{c}{\textbf{6.75}}\\
\hhline{=========}
\end{tabular}
} 
\caption{Results from 40 real observational datasets. Lower MISE loss values (up to a constant) denote better performance. Underlined values correspond to the best in each row. The second column lists the rank of DDR relative to the other algorithms. DDR performs the best on average by achieving the lowest average rank as shown in the last row.} \label{table_real}
\end{table*}

\begin{table*}
\footnotesize
\centerline{
\begin{tabular}{cccccccccc}
\hhline{==========}
                  & $T$ & (rank) & DDR  & FC-KRR & FC-NN & FC-Lasso & SCDE & LSCDE & NFE  \\ \hline
CSP-1019          & 0          & 2               & \Minus9.69E\Minus02     & \Minus8.24E\Minus02       & \Minus9.50E\Minus02      & \Minus8.49E\Minus02         & \underline{\Minus1.00E\Minus01}     & \Minus8.64E\Minus02      & \Plus4.59E\Minus02      \\
                  & 1          & 1               & \underline{\Minus9.91E\Minus02}     & \Minus9.77E\Minus02       & \Minus9.37E\Minus02      & \Minus4.92E\Minus02         & \Minus6.28E\Minus02     & \Minus9.51E\Minus02      & \Plus1.47E\Minus01      \\
CSP-1021          & 0          & 1               & \underline{\Minus1.09E\Minus01}     & \Minus8.29E\Minus02       & \Minus9.78E\Minus02      & \Plus1.52E\Minus03          & \Plus3.08E\Minus01      & \Minus7.35E\Minus02      & \Plus2.10E\Plus00      \\
                  & 1          & 2               & \Minus1.06E\Minus01     & \Minus7.76E\Minus02       & \Minus9.40E\Minus02      & \Minus7.68E\Minus02         & \Minus6.78E\Minus02     & \underline{\Minus1.10E\Minus01}      & \Plus1.52E\Plus00      \\ 
CTN-0009          & 0          & 2               & \Minus1.66E\Minus02     & \Minus1.39E\Minus02       & \Minus1.21E\Minus02      & \Minus1.93E\Minus04         & \Minus2.36E\Minus03     & \underline{\Minus1.83E\Minus02}      & \Plus1.78E\Plus01      \\
                  & 1          & 2               & \Minus1.82E\Minus02     & \Minus1.79E\Minus02       & \Minus1.80E\Minus02      & \Minus9.56E\Minus03         & \Minus1.26E\Minus02     & \underline{\Minus1.94E\Minus02}      & \Plus7.50E\Plus00      \\ 
CTN-0010          & 0          & 1               & \underline{\Minus1.01E\Minus02}     & \Minus5.93E\Minus03       & \Minus1.00E\Minus02      & \Plus2.72E\Minus03          & \Plus2.96E\Minus02      & \Minus3.80E\Minus03      & \Plus4.65E\Plus02      \\
                  & 1          & 1               & \underline{\Minus1.37E\Minus02}     & \Plus3.95E\Minus03        & \Minus1.26E\Minus02      & \Plus2.37E\Minus02          & \Minus2.72E\Minus03     & \Minus1.13E\Minus02      & \Plus4.72E\Plus01      \\ 
CTO-0012          & 0          & 4               & \Minus7.08E\Minus02     & \Minus7.52E\Minus02       & \underline{\Minus8.89E\Minus02}      & \Minus2.31E\Minus02         & 3.91E\Minus02      & \Minus7.20E\Minus02      & \Plus2.10E\Plus01      \\
                  & 1          & 1               & \underline{\Minus9.96E\Minus02}     & \Minus8.97E\Minus02       & \Minus7.60E\Minus02      & \Plus1.38E\Minus03          & \Plus1.72E\Minus01      & \Minus4.19E\Minus02      & \Plus9.58E\Plus00      \\ 
CSP-1025          & 0          & 3               & \Minus4.36E\Minus01     & \Minus1.39E\Minus01       & \Minus2.74E\Minus01      & \Minus8.64E\Minus01         & \underline{\Minus1.13E\Plus00}     & \Minus1.34E\Minus01      & \Plus5.40E\Plus00      \\
                  & 1          & 4               & \Minus2.46E\Minus01     & \Minus2.86E\Minus01       & \Minus1.49E\Minus01      & \underline{\Minus7.16E\Minus01}         & \Minus5.63E\Minus01     & \Minus6.96E\Minus02      & \Plus1.69E\Plus00      \\  
CTN-0051          & 0          & 4               & \Minus6.49E\Minus02     & \Minus1.36E\Minus01       & \Minus9.57E\Minus02      & \underline{\Minus1.42E\Minus01}         & 9.43E\Minus03      & \Minus3.72E\Minus02      & \Plus2.83E\Plus01      \\
                  & 1          & 4               & \Minus9.38E\Minus02     & \Minus2.11E\Minus01       & \Minus2.03E\Minus01      & \underline{\Minus2.36E\Minus01}         & \Minus5.85E\Minus02     & \Minus3.97E\Minus02      & \Plus1.00E\Plus01      \\ \hline
\textbf{Avg Rank} & \textbf{}  & \textbf{}       & \textbf{2.29} & \textbf{3.29}   & \textbf{3.00}  & \textbf{4.21}     & \textbf{4.57} & \textbf{3.64}  & \textbf{7.00}\\
\hhline{==========}
\end{tabular}
}
\caption{Results from 7 real clinical trials presented in a similar format as Table \ref{table_real}. DDR again achieves the lowest average rank.} \label{table_CT}
\end{table*}

\subsection{Detailed Clinical Trial Application} \label{sec_CT}

We now provide a detailed application of DDR to real RCT data in order to recover conditional densities summarizing treatment outcome. The discovery reported below is novel even in the medical literature. We investigated the effect of transdermal nicotine patches (TNPs) on long-term smoking cessation using the CTN-0009 dataset from the NIDA Data Share \citep{Reid08}. In this trial, 166 subjects were randomized to receive either TNP or treatment-as-usual (TAU; motivational interviewing and supportive therapy) without TNP over a period of 8 weeks. Smoking increases carbon monoxide (CO) levels in the lungs, so the investigators objectively monitored smoking cessation by measuring CO levels with a breathalyzer.

\begin{figure}
    \centering
    \begin{subfigure}[b]{0.32\textwidth}
    \centering
     \includegraphics[scale=0.45]{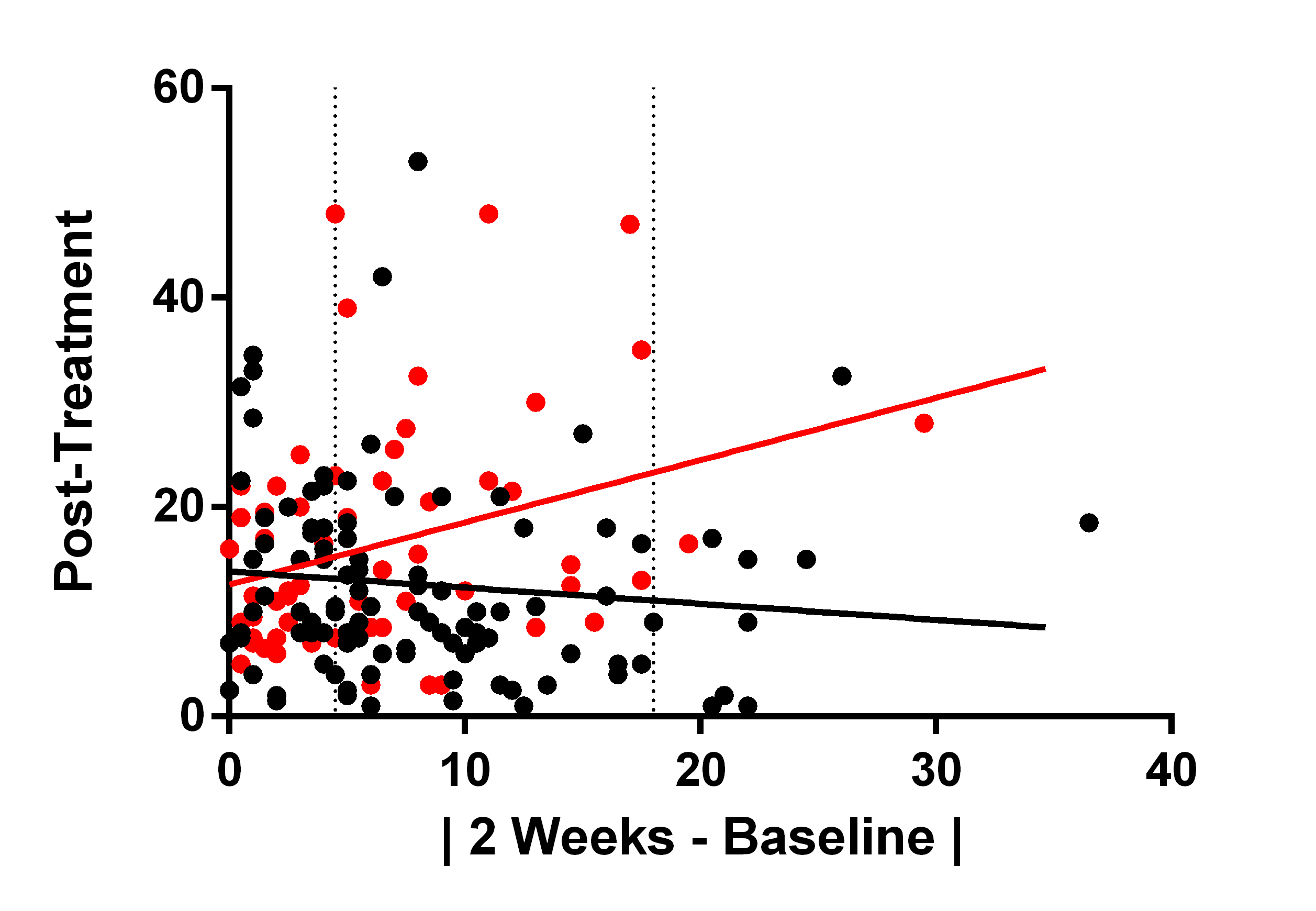}
        \caption{} \label{fig_nicotine:diff}
    \end{subfigure}
    \begin{subfigure}[b]{0.32\textwidth}
    \centering
     \includegraphics[scale=0.45]{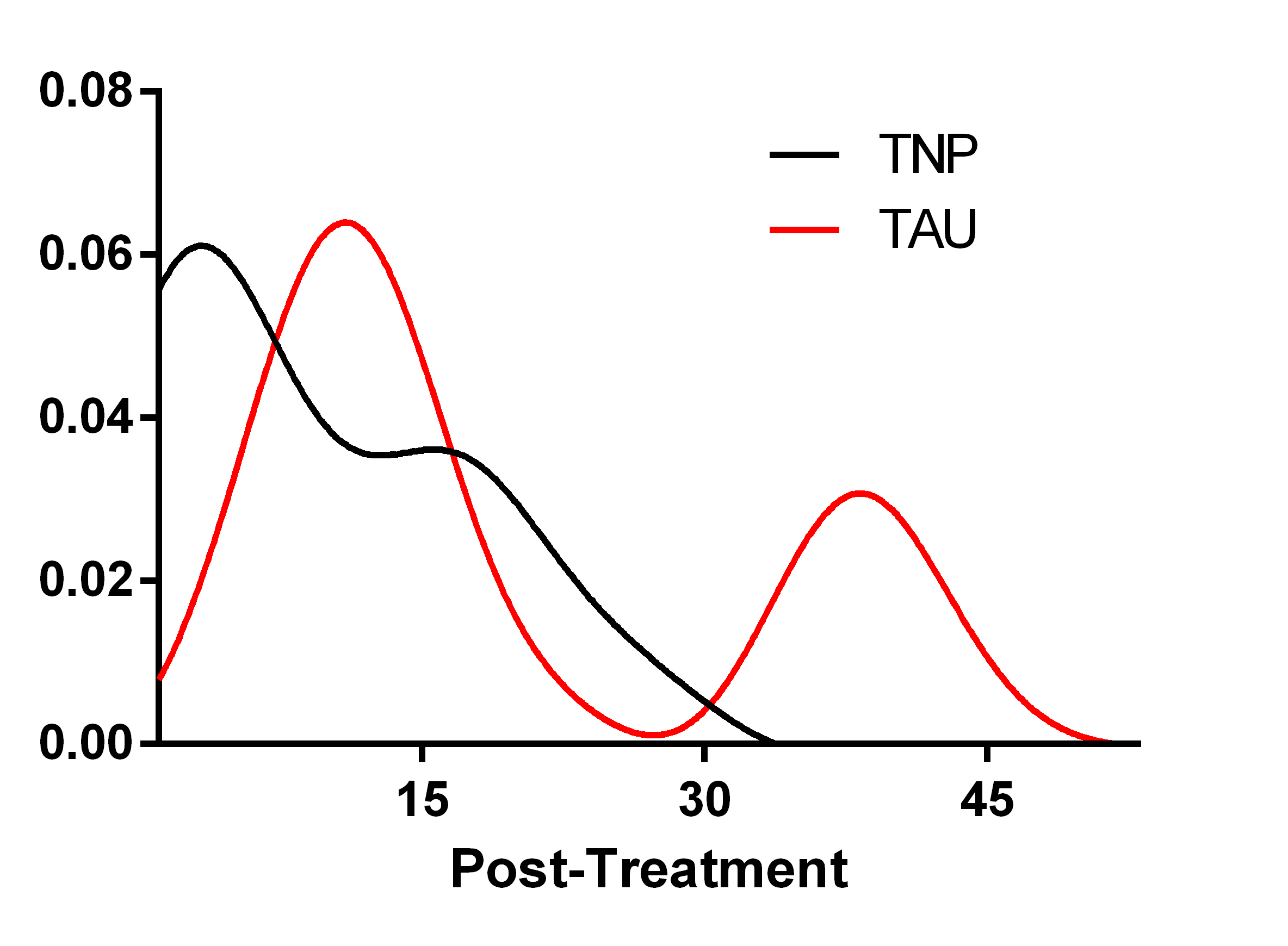}
     \caption{}\label{fig_nicotine:pt1}
    \end{subfigure}
     \begin{subfigure}[b]{0.32\textwidth}
     \centering
     \includegraphics[scale=0.453]{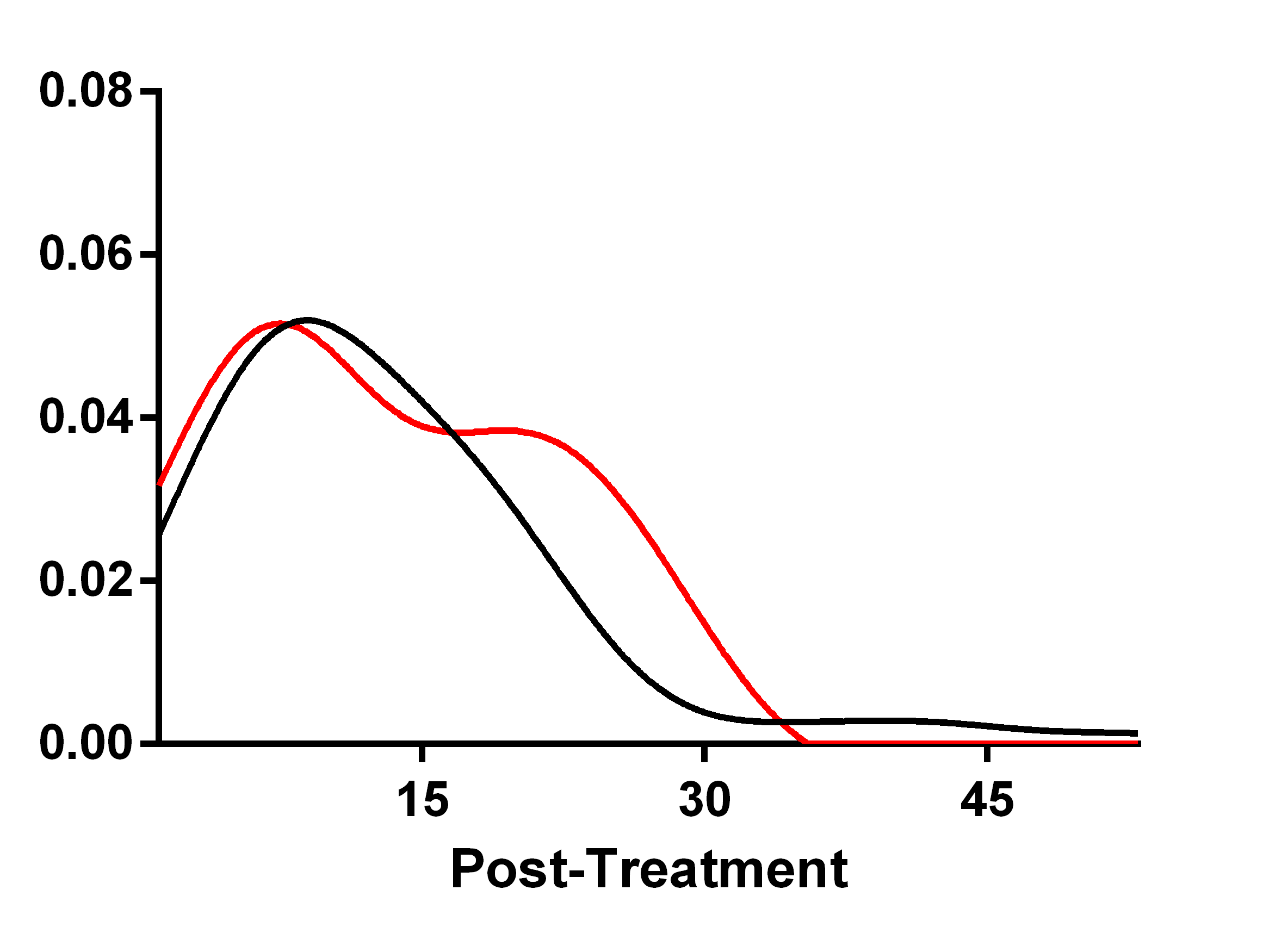}
        \caption{}\label{fig_nicotine:pt2}
    \end{subfigure}
    \caption{Detailed analysis of a clinical trial dataset for personalized outcomes of TNP. (a) We can summarize the personalized outcomes of TNP relative to TAU using two linear regression slopes. (b) The CO levels of this particular patient changed by 18 ppm, so DDR predicts that TNP is more beneficial to this patient than TAU. (c) On the other hand, CO levels only changed by 4.5 ppm in this patient, so DDR could not clearly differentiate the outcomes of the two treatments.}
    \label{fig_nicotine}
\end{figure}

TNPs only mildly increase smoking cessation after treatment ends (see Appendix \ref{sec_ex_CT}). The small effect size may exist because only a minority of patients benefit from TNP. We in particular hypothesized that patients who smoke irregularly experience more intense nicotine cravings than those who smoke on a regular basis. Patients who smoke irregularly may want to stop smoking, but they struggle to remain in remission due to the cravings. As a result, these individuals will benefit more from TNP because TNP decreases the frequency and intensity of the craving episodes \citep{Rose85}. We can evaluate this hypothesis using the RCT dataset, where we track the consistency in smoking using short-term changes in lung CO levels as shown on the x-axis in Figure \ref{fig_nicotine:diff}. The x-axis more specifically corresponds to the absolute value of the CO levels at baseline minus the CO levels at 2 weeks. The y-axis denotes the CO levels at 9 weeks after treatment ended. Based on the two different linear regression slopes, we can see that a change in CO levels while on TNP generally has no effect on post-treatment CO levels but a change in CO levels while on TAU has a detrimental effect.\footnote{Non-linear polynomial regression produced essentially the same conditional expectation estimates.} We confirmed the significance of the observed trend by rejecting the null of equality in slopes between TNP and TAU (z=-2.87, one sided p=0.002). We therefore conclude that TNP reduces post-treatment CO levels among patients with large changes in short term CO levels.

The regression slopes however only provide point estimates of predicted treatment outcome. In reality, even patients with large changes may not benefit from TNP because the post-treatment CO level is stochastic. DDR allows us to visualize this uncertainty by recovering  conditional densities for all possible outcome values. Consider for example patient A with a value of 18 on the x-axis of Figure \ref{fig_nicotine:diff}; the regression slopes suggest that this patient is expected to obtain a post-treatment CO level of around 11 if treated with TNP but a level of 23 if treated with TAU. The densities recovered by DDR in Figure  \ref{fig_nicotine:pt1} for patient A are also significantly different (one sided p=0.013; see Appendix \ref{sec_hypothesis}), but they imply a more modest effect because the patient also has a high probability of \textit{not} experiencing such a large difference in post-treatment CO levels. On the other hand, patient B with a value of 4.5 on the x-axis is predicted to respond equally well to TNP and TAU in Figure \ref{fig_nicotine:diff}; the patient may therefore decide not to take TNP based on the regression estimates. However, DDR recovers broad densities as shown in Figure \ref{fig_nicotine:pt2}; these densities imply that we cannot differentiate between the two treatments with the available evidence, so patient B may actually benefit more from TNP than TAU. This patient may therefore decide to try TNP after seeing the output of DDR. We conclude that the conditional densities recovered by DDR help patients make more informed treatment decisions by allowing them to visualize the probabilities associated with all possible outcome values in an intuitive fashion.

\section{Conclusion} \label{sec_concl}

We proposed DDR for estimating conditional densities by performing non-linear regression over a set of kernel density functions. DDR differs from CKDE by directly estimating the conditional density, as opposed to estimating the joint and marginal densities first. DDR outperforms previous methods on average across a variety of synthetic and real datasets. The algorithm also generates patient-specific densities of treatment outcomes when run on RCT data; as opposed to the conditional expectation, the conditional density recovered by DDR allows patients and healthcare providers to easily compare the probabilities associated with \textit{all} possible outcome values by effectively converting a standard RCT into a personalized one. Theoretical results further support our empirical claims by highlighting the consistency of DDR as well as quantifying the rate of bias with respect to the smoothing parameter $h$. We ultimately believe that this work is an important contribution to the literature because it introduces a state of the art conditional density estimation method as well as demonstrates a non-trivial application to a key problem in medicine.

\bibliography{bibliog.bib}

\begin{thebibliography}{39}
\providecommand{\natexlab}[1]{#1}
\providecommand{\url}[1]{\texttt{#1}}
\expandafter\ifx\csname urlstyle\endcsname\relax
  \providecommand{\doi}[1]{doi: #1}\else
  \providecommand{\doi}{doi: \begingroup \urlstyle{rm}\Url}\fi

\bibitem[Akobeng(2007)]{Akobeng07}
Anthony~K Akobeng.
\newblock Understanding diagnostic tests 2: likelihood ratios, pre-and
  post-test probabilities and their use in clinical practice.
\newblock \emph{Acta Paediatrica}, 96\penalty0 (4):\penalty0 487--491, 2007.

\bibitem[Berger(2015)]{Berger15}
Zackary Berger.
\newblock Navigating the unknown: shared decision-making in the face of
  uncertainty.
\newblock \emph{Journal of General Internal Medicine}, 30\penalty0
  (5):\penalty0 675--678, 2015.

\bibitem[Bhise et~al.(2018)Bhise, Meyer, Menon, Singhal, Street, Giardina, and
  Singh]{Bhise18}
Viraj Bhise, Ashley~ND Meyer, Shailaja Menon, Geeta Singhal, Richard~L Street,
  Traber~D Giardina, and Hardeep Singh.
\newblock Patient perspectives on how physicians communicate diagnostic
  uncertainty: an experimental vignette study.
\newblock \emph{International Journal for Quality in Health Care}, 30\penalty0
  (1):\penalty0 2--8, 2018.

\bibitem[Bishop(2006)]{Bishop06}
Christopher~M Bishop.
\newblock \emph{Pattern Recognition and Machine Learning}.
\newblock Springer, 2006.

\bibitem[Cattaneo et~al.(2018)Cattaneo, Jansson, and Ma]{Cattaneo18}
Matias~D Cattaneo, Michael Jansson, and Xinwei Ma.
\newblock Simple local polynomial density estimators.
\newblock \emph{arXiv preprint arXiv:1811.11512}, 2018.

\bibitem[Cruz-Uribe and Neugebauer(2003)]{Cruz03}
David Cruz-Uribe and CJ~Neugebauer.
\newblock An elementary proof of error estimates for the trapezoidal rule.
\newblock \emph{Mathematics Magazine}, 76\penalty0 (4):\penalty0 303--306,
  2003.

\bibitem[Diamond and Forrester(1979)]{Diamond79}
George~A Diamond and James~S Forrester.
\newblock Analysis of probability as an aid in the clinical diagnosis of
  coronary-artery disease.
\newblock \emph{New England Journal of Medicine}, 300\penalty0 (24):\penalty0
  1350--1358, 1979.

\bibitem[Fan et~al.(1996)Fan, Yao, and Tong]{Fan96}
Jianqing Fan, Qiwei Yao, and Howell Tong.
\newblock Estimation of conditional densities and sensitivity measures in
  nonlinear dynamical systems.
\newblock \emph{Biometrika}, 83\penalty0 (1):\penalty0 189--206, 1996.

\bibitem[Fosgerau and Fukuda(2010)]{Fosgerau10}
Mogens Fosgerau and Daisuke Fukuda.
\newblock {Valuing travel time variability: Characteristics of the travel time
  distribution on an urban road}.
\newblock MPRA Paper 24330, University Library of Munich, Germany, 2010.
\newblock URL \url{https://ideas.repec.org/p/pra/mprapa/24330.html}.

\bibitem[Fryer(1976)]{Fryer76}
MJ~Fryer.
\newblock Some errors associated with the non-parametric estimation of density
  functions.
\newblock \emph{IMA Journal of Applied Mathematics}, 18\penalty0 (3):\penalty0
  371--380, 1976.

\bibitem[Hyndman et~al.(1996)Hyndman, Bashtannyk, and Grunwald]{Hyndman96}
Rob~J. Hyndman, David~M. Bashtannyk, and Gary~K. Grunwald.
\newblock Estimating and visualizing conditional densities.
\newblock \emph{Journal of Computational and Graphical Statistics}, 5\penalty0
  (4):\penalty0 315--336, 1996.
\newblock \doi{10.1080/10618600.1996.10474715}.
\newblock URL
  \url{https://amstat.tandfonline.com/doi/abs/10.1080/10618600.1996.10474715}.

\bibitem[Izbicki and Lee(2017)]{Izbicki17}
Rafael Izbicki and Ann Lee.
\newblock Converting high-dimensional regression to high-dimensional
  conditional density estimation.
\newblock \emph{Electron. J. Statist.}, 11\penalty0 (2):\penalty0 2800--2831,
  2017.
\newblock \doi{10.1214/17-EJS1302}.
\newblock URL \url{https://doi.org/10.1214/17-EJS1302}.

\bibitem[Izbicki and Lee(2016)]{Izbicki16}
Rafael Izbicki and Ann~B. Lee.
\newblock Nonparametric conditional density estimation in a high-dimensional
  regression setting.
\newblock \emph{Journal of Computational and Graphical Statistics}, 25\penalty0
  (4):\penalty0 1297--1316, 2016.
\newblock \doi{10.1080/10618600.2015.1094393}.
\newblock URL \url{https://doi.org/10.1080/10618600.2015.1094393}.

\bibitem[Izmailov et~al.(2013)Izmailov, Vapnik, and Vashist]{Izmailov13}
Rauf Izmailov, Vladimir Vapnik, and Akshay Vashist.
\newblock Multidimensional splines with infinite number of knots as svm
  kernels.
\newblock In \emph{The 2013 International Joint Conference on Neural Networks
  (IJCNN)}, pages 1--7. IEEE, 2013.

\bibitem[Kanamori et~al.(2012)Kanamori, Suzuki, and Sugiyama]{Kanamori12}
Takafumi Kanamori, Taiji Suzuki, and Masashi Sugiyama.
\newblock Statistical analysis of kernel-based least-squares density-ratio
  estimation.
\newblock \emph{Machine Learning}, 86\penalty0 (3):\penalty0 335--367, 2012.

\bibitem[Kanwal(2011)]{Kanwal11}
Ram~P Kanwal.
\newblock \emph{Generalized Functions: Theory and Applications}.
\newblock Springer Science \& Business Media, 2011.

\bibitem[Luedtke and van~der Laan(2016)]{Luedtke16}
Alexander~R. Luedtke and Mark~J. van~der Laan.
\newblock Statistical inference for the mean outcome under a possibly
  non-unique optimal treatment strategy.
\newblock \emph{Ann. Statist.}, 44\penalty0 (2):\penalty0 713--742, 04 2016.
\newblock \doi{10.1214/15-AOS1384}.
\newblock URL \url{https://doi.org/10.1214/15-AOS1384}.

\bibitem[Mah et~al.(2016)Mah, Muthupalaniappen, and Chong]{Mah16}
Hui~Chin Mah, Leelavathi Muthupalaniappen, and Wei~Wen Chong.
\newblock Perceived involvement and preferences in shared decision-making among
  patients with hypertension.
\newblock \emph{Family Practice}, 33\penalty0 (3):\penalty0 296--301, 2016.

\bibitem[Martinkovich et~al.(2014)Martinkovich, Shah, Planey, and
  Arnott]{Martinkovich14}
Stephen Martinkovich, Darshan Shah, Sonia~Lobo Planey, and John~A Arnott.
\newblock Selective estrogen receptor modulators: tissue specificity and
  clinical utility.
\newblock \emph{Clinical Interventions in Aging}, 9:\penalty0 1437, 2014.

\bibitem[Newey and McFadden(1994)]{Newey94}
Whitney~K Newey and Daniel McFadden.
\newblock Large sample estimation and hypothesis testing.
\newblock \emph{Handbook of Econometrics}, 4:\penalty0 2111--2245, 1994.

\bibitem[Reid et~al.(2008)Reid, Fallon, Sonne, Flammino, Nunes, Jiang,
  Kourniotis, Lima, Brady, Burgess, Arfken, Pihlgren, Giordano, Starosta,
  Robinson, and Rotrosen]{Reid08}
M.~S. Reid, B.~Fallon, S.~Sonne, F.~Flammino, E.~V. Nunes, H.~Jiang,
  E.~Kourniotis, J.~Lima, R.~Brady, C.~Burgess, C.~Arfken, E.~Pihlgren,
  L.~Giordano, A.~Starosta, J.~Robinson, and J.~Rotrosen.
\newblock {{S}moking cessation treatment in community-based substance abuse
  rehabilitation programs}.
\newblock \emph{J Subst Abuse Treat}, 35\penalty0 (1):\penalty0 68--77, Jul
  2008.

\bibitem[Rose et~al.(1985)Rose, Herskovic, Trilling, and Jarvik]{Rose85}
Jed~E Rose, Joseph~E Herskovic, Yvonne Trilling, and Murray~E Jarvik.
\newblock Transdermal nicotine reduces cigarette craving and nicotine
  preference.
\newblock \emph{Clinical Pharmacology \& Therapeutics}, 38\penalty0
  (4):\penalty0 450--456, 1985.

\bibitem[Rosenbaum and Rubin(1983)]{Rosenbaum83}
Paul~R Rosenbaum and Donald~B Rubin.
\newblock The central role of the propensity score in observational studies for
  causal effects.
\newblock \emph{Biometrika}, 70\penalty0 (1):\penalty0 41--55, 1983.

\bibitem[Rosenblatt(1969)]{Rosenblatt69}
M~Rosenblatt.
\newblock Conditional probability density and regression estimators.
\newblock \emph{International Symposium on Multivariate Analysis, Multivariate
  analysis II Proceedings}, 1969.

\bibitem[Rubin(1974)]{Rubin74}
Donald~B Rubin.
\newblock Estimating causal effects of treatments in randomized and
  nonrandomized studies.
\newblock \emph{Journal of Educational Psychology}, 66\penalty0 (5):\penalty0
  688, 1974.

\bibitem[Rubin(1990)]{Rubin90}
Donald~B Rubin.
\newblock [on the application of probability theory to agricultural
  experiments. essay on principles. section 9.] comment: Neyman (1923) and
  causal inference in experiments and observational studies.
\newblock \emph{Statistical Science}, 5\penalty0 (4):\penalty0 472--480, 1990.

\bibitem[Seber and Lee(2012)]{Seber12}
G.A.F. Seber and A.J. Lee.
\newblock \emph{Linear Regression Analysis}.
\newblock Wiley Series in Probability and Statistics. Wiley, 2012.
\newblock ISBN 9781118274422.
\newblock URL \url{https://books.google.com/books?id=X2Y6OkXl8ysC}.

\bibitem[Shiga et~al.(2015)Shiga, Tangkaratt, and Sugiyama]{Shiga15}
Motoki Shiga, Voot Tangkaratt, and Masashi Sugiyama.
\newblock Direct conditional probability density estimation with sparse feature
  selection.
\newblock \emph{Machine Learning}, 100\penalty0 (2):\penalty0 161--182, Sep
  2015.
\newblock ISSN 1573-0565.
\newblock \doi{10.1007/s10994-014-5472-x}.
\newblock URL \url{https://doi.org/10.1007/s10994-014-5472-x}.

\bibitem[Smale and Zhou(2007)]{Smale07}
Steve Smale and Ding-Xuan Zhou.
\newblock Learning theory estimates via integral operators and their
  approximations.
\newblock \emph{Constructive Approximation}, 26\penalty0 (2):\penalty0
  153--172, Aug 2007.
\newblock ISSN 1432-0940.
\newblock \doi{10.1007/s00365-006-0659-y}.
\newblock URL \url{https://doi.org/10.1007/s00365-006-0659-y}.

\bibitem[Sugiyama et~al.(2010)Sugiyama, Takeuchi, Suzuki, Kanamori, Hachiya,
  and Okanohara]{Sugiyama10}
Masashi Sugiyama, Ichiro Takeuchi, Taiji Suzuki, Takafumi Kanamori, Hirotaka
  Hachiya, and Daisuke Okanohara.
\newblock Conditional density estimation via least-squares density ratio
  estimation.
\newblock In Yee~Whye Teh and Mike Titterington, editors, \emph{Proceedings of
  the Thirteenth International Conference on Artificial Intelligence and
  Statistics}, volume~9 of \emph{Proceedings of Machine Learning Research},
  pages 781--788, Chia Laguna Resort, Sardinia, Italy, 13--15 May 2010. PMLR.
\newblock URL \url{http://proceedings.mlr.press/v9/sugiyama10a.html}.

\bibitem[Takeuchi et~al.(2006)Takeuchi, Le, Sears, and Smola]{Takeuchi06}
Ichiro Takeuchi, Quoc~V. Le, Timothy~D. Sears, and Alexander~J. Smola.
\newblock Nonparametric quantile estimation.
\newblock \emph{J. Mach. Learn. Res.}, 7:\penalty0 1231--1264, December 2006.
\newblock ISSN 1532-4435.
\newblock URL \url{http://dl.acm.org/citation.cfm?id=1248547.1248592}.

\bibitem[Tangkaratt et~al.(2015)Tangkaratt, Xie, and Sugiyama]{Tangkaratt15}
Voot Tangkaratt, Ning Xie, and Masashi Sugiyama.
\newblock Conditional density estimation with dimensionality reduction via
  squared-loss conditional entropy minimization.
\newblock \emph{Neural Computation}, 27:\penalty0 228--254, 2015.

\bibitem[Tresp(2001)]{Tresp01}
Volker Tresp.
\newblock Mixtures of gaussian processes.
\newblock In \emph{Advances in Neural Information Processing Systems}, pages
  654--660, 2001.

\bibitem[Trippe and Turner(2018)]{Trippe18}
Brian~L Trippe and Richard~E Turner.
\newblock Conditional density estimation with bayesian normalising flows.
\newblock \emph{arXiv preprint arXiv:1802.04908}, 2018.

\bibitem[Vaart(1998)]{Vaart98}
A.~W. van~der Vaart.
\newblock \emph{Asymptotic Statistics}.
\newblock Cambridge Series in Statistical and Probabilistic Mathematics.
  Cambridge University Press, 1998.
\newblock \doi{10.1017/CBO9780511802256}.

\bibitem[Vapnik(2013)]{Vapnik13}
Vladimir Vapnik.
\newblock \emph{The Nature of Statistical Learning Theory}.
\newblock Springer science \& business media, 2013.

\bibitem[Wand and Jones(1994)]{Wand94}
Matt~P Wand and M~Chris Jones.
\newblock \emph{Kernel smoothing}.
\newblock Chapman and Hall/CRC, 1994.

\bibitem[Zhang et~al.(2012{\natexlab{a}})Zhang, Tsiatis, Davidian, Zhang, and
  Laber]{Zhang2012c}
Baqun Zhang, Anastasios~A Tsiatis, Marie Davidian, Min Zhang, and Eric Laber.
\newblock Estimating optimal treatment regimes from a classification
  perspective.
\newblock \emph{Stat}, 1\penalty0 (1):\penalty0 103--114, 2012{\natexlab{a}}.

\bibitem[Zhang et~al.(2012{\natexlab{b}})Zhang, Tsiatis, Laber, and
  Davidian]{Zhang12}
Baqun Zhang, Anastasios~A Tsiatis, Eric~B Laber, and Marie Davidian.
\newblock A robust method for estimating optimal treatment regimes.
\newblock \emph{Biometrics}, 68\penalty0 (4):\penalty0 1010--1018,
  2012{\natexlab{b}}.

\end{thebibliography}

\section{Appendix}

\subsection{Proofs} \label{sec_proofs}
\begin{definition1}
(Stochastic equicontinuity w.r.t. $z$) For every $\varepsilon, \delta>0$, there exists a sequence of random variables $\Upsilon_n$ and an integer $N$ such that $\forall n \geq N$, we have $\mathbb{P}(|\Upsilon_n|>\varepsilon) < \delta$. Moreover, for each $z$, there is an open set $\mathcal{N}$ containing $z$ with:
$$\sup_{z^\prime \in \mathcal{N}} |Q_n(z) - Q_n(z^\prime)| \leq \Upsilon_n, ~~ n \geq N.$$
\end{definition1}
\noindent Notice that $\Upsilon_n$ acts like a random epsilon by bounding changes in $Q_n(z)$ w.r.t. $z$.

\begin{lemma1} \label{lem_lipschitz}
(Lipschitz continuity $\implies$ stochastic equicontinuity; Lemma 2.9 in \citep{Newey94})  If $Q_n(z) = o_p(1)$ for all $z \in [a,b]$ and $B_n=O_p(1)$ such that for all $z,z^\prime \in [a,b]$ we have $| Q_n(z) - Q_n(z^\prime)| \leq B_n |z-z^\prime |$, then $Q_n(z)$ is stochastically equicontinuous w.r.t. $z$.
\end{lemma1}

\begin{lemma1}  \label{lem_differentiable}
If Assumption 2 holds, then $Q_n(z)$ is stochastically equicontinuous w.r.t. $z$.
\end{lemma1}
\begin{proof}
Because $Q_n(z)$ is differentiable on $[a,b]$ w.r.t. $z$, we can write:
$$ | Q_n(z) - Q_n(z^\prime)| \leq \sup_{z \in [a,b]} \Big| \frac{\partial Q_n(z)}{\partial z} \Big||z-z^\prime |.$$
We have $\sup_{z \in [a,b]} \Big| \frac{\partial Q_n(z)}{\partial z} \Big|=O_p(1)$ by Assumption 2. Invoke Lemma \ref{lem_lipschitz} to conclude that $Q_n$ is stochastically equicontinuous w.r.t. $z$.
\end{proof}

\begin{lemma1}  \label{lem_uniform}
(Stochastic equicontinuity + pointwise consistency $\iff$ uniform consistency; Lemma 2.8 in \citep{Newey94}) We have $Q_n(z) = o_p(1)$ for all $z \in [a,b]$ and $Q_n(z)$ is stochastically equicontinuous w.r.t $z$ if and only if $\sup_{z \in [a,b]} Q_n(z) = o_p(1)$.
\end{lemma1}

\begin{lemma1}  \label{lem_integral}
(Uniform consistency $\implies$ proper integral consistency) If we have $\sup_{z \in [a,b]}$ $Q_n(z) = o_p(1)$, then $\int_a^b Q_n(z) ~dz = o_p(1)$.
\end{lemma1}
\begin{proof}
Choose $\varepsilon > 0$. Then write:
\begin{equation}
    \begin{aligned}
    &\hspace{4.75mm}\mathbb{P}\Big( \Big|\int_a^b Q_n(z) ~dz\Big| \geq \varepsilon\Big)\\ &\leq \mathbb{P}\Big( \int_a^b |Q_n(z)| ~dz \geq \varepsilon \Big)\\
    &\leq \mathbb{P}\Big( \int_a^b \sup_{z \in [a,b]}|Q_n(z)| ~dz \geq \varepsilon\Big)\\
    &=\mathbb{P}\Big( (b-a)\sup_{z \in [a,b]}|Q_n(z)| \geq \varepsilon \Big)\\
    &=\mathbb{P}\Big( \sup_{z \in [a,b]}|Q_n(z)| \geq \frac{\varepsilon}{b-a}\Big).
    \end{aligned}
\end{equation}
Choose $\delta>0$. By assumption, for $\frac{\varepsilon}{b-a}$ and $\delta$, $\exists N \in \mathbb{N}^+$ such that $\forall n \geq N$, we have:
$$\mathbb{P}\Big( \sup_{z \in [a,b]}|Q_n(z)| \geq \frac{\varepsilon}{b-a}\Big) \leq \delta.$$ Note that we chose $\varepsilon$ and $\delta$ arbitrarily. The conclusion follows by the epsilon-delta definition of convergence in probability to zero.
\end{proof}

\begin{customthm}{1}
Under Assumptions 1-4, we have:
\begin{equation} \nonumber
\int_a^b \int \Big|\widehat{g}_{h}(z|\bm{x}) - f_{Y(t)}(z |\bm{x})\Big|^2 ~ d \mathbb{P}_{\bm{X}}(\bm{x})dz \leq o_p(1) + Ch^4,
\end{equation}
for any $a<b$ where $C$ is a constant that does not depend on $n$ or $h$.
\end{customthm}
\begin{proof}
We write:
\begin{equation} \label{eq_thm1:bound}
\begin{aligned}
&\int_a^b \int \Big|\widehat{g}_{h}(z|\bm{x}) - f_{Y(t)}(z|\bm{x}) \Big|^2 ~ d \mathbb{P}_{\bm{X}}(\bm{x}) dz\\
\leq \hspace{1mm}&2\int_a^b\int \Big|\widehat{g}_{h}(z|\bm{x}) - g_{h} (z|\bm{x}) \Big|^2 ~ d \mathbb{P}_{\bm{X}}(\bm{x})dz\\ + &2\int_a^b\int \Big|g_{h} (z|\bm{x}) - f_{Y(t)}(z |\bm{x}) \Big|^2 ~ d \mathbb{P}_{\bm{X}}(\bm{x})dz
\end{aligned}
\end{equation}
The term $Q_n(z) =\int \Big|\widehat{g}_{h}(z|\bm{x}) - g_{h} (z|\bm{x}) \Big|^2 ~ d \mathbb{P}_{\bm{X}}(\bm{x})$ is $o_p(1)$ for each $z \in [a,b]$ by Assumption 1. Invoke Lemmas 3, 4 and then 5 to conclude that we have:
\begin{equation} \nonumber
    \int_a^b\int \Big|\widehat{g}_{h}(z|\bm{x}) - g_{h} (z|\bm{x}) \Big|^2 ~ d \mathbb{P}_{\bm{X}}(\bm{x})dz = o_p(1).
\end{equation}

We now focus on the term $ \Big|g_{h} (z|\bm{x}) - f_{Y(t)}(z |\bm{x}) \Big|^2$. We write:
\begin{equation} \label{eq_thm1:kernel}
\begin{aligned}
&\hspace{4.75mm}g_{h}(z|\bm{x}) - f_{Y(t)}(z|\bm{x})\\ &= \int \frac{1}{h}K\Big( \frac{z-k}{h}\Big) f_{Y(t)}(k|\bm{x}) ~dk - f_{Y(t)}(z|\bm{x})\\
&=\int K(u) \Big( f_{Y(t)}(z - hu | \bm{x}) - f_{Y(t)}(z |\bm{x}) \Big) ~du.
\end{aligned}
\end{equation}
We then utilize a Taylorian expansion with a Laplacian representation of the remainder:
\begin{equation} \nonumber
\begin{aligned}
    &\hspace{5mm}f_{Y(t)}(z + h|\bm{x}) - f_{Y(t)}(z|\bm{x})\\
    &= hf_{Y(t)}'(z|\bm{x}) + h^2 \int_0^1 f_{Y(t)}''(z+sh|\bm{x})(1-s)~ds.
\end{aligned}
\end{equation}
Substituting the above formula into Equation \eqref{eq_thm1:kernel}, we get:
\begin{equation} \label{eq_thm1:taylor}
\begin{aligned}
&\hspace{5.5mm} g_{h} (z|\bm{x}) - f_{Y(t)}(z|\bm{x})\\ &= \int \int_0^1 K(u) [ -huf_{Y(t)}'(z|\bm{x})+h^2u^2 f_{Y(t)}''(z-shu|\bm{x})(1-s)]~dsdu\\
&= \int \int_0^1 h^2uK(u) \Big(u f_{Y(t)}''(z- shu|\bm{x})(1-s)\Big)~dsdu,
\end{aligned}
\end{equation}
where the second equality follows because we assumed that $K$ has expectation zero. We next utilize the Cauchy-Schwartz inequality $(\mathbb{E}AB)^2 \leq \mathbb{E}A^2\mathbb{E}B^2$ with $A = U$ and $B=Uf''(z-ShU|\bm{x})(1-S)$; here $U$ has density $K$ and $S$ is uniform on $[0,1]$ as well as independent of $U$. The bottom of Equation \eqref{eq_thm1:taylor} squared is therefore upper bounded by:
\begin{equation} \nonumber
\begin{aligned}
    &h^4 \int K(u)u^2 ~du \int \int_0^1 K(u)u^2 f_{Y(t)}''(z-shu|\bm{x})^2(1-s)^2 ~dsdu\\
    =\hspace{1mm}&h^4 \int K(u)u^2 ~du \int \int_0^1 K(u)u^2 f_{Y(t)}''(z|\bm{x})^2(1-s)^2 ~dsdu\\
    =\hspace{1mm}&h^4 \Big(\int K(u)u^2 ~du \Big)^2 f_{Y(t)}''(z|\bm{x})^2 \frac{1}{3}.
\end{aligned}
\end{equation}
Integrating this with respect to $\mathbb{P}_{\bm{X}}(\bm{x})$ and $z$, we obtain:
\begin{equation} \nonumber
\begin{aligned}
    &\int_a^b \int \Big( g_{h} (z|\bm{x}) - f_{Y(t)}(z|\bm{x}) \Big)^2 ~ d \mathbb{P}_{\bm{X}}(\bm{x})dz\\ \leq \hspace{1mm}&h^4 \Big(\int K(u)u^2 ~du \Big)^2 \frac{1}{3}\int_a^b \int f_{Y(t)}''(z|\bm{x})^2 ~ d \mathbb{P}_{\bm{X}}(\bm{x})dz.
\end{aligned}
\end{equation}
We finally utilize the bound in Equation \eqref{eq_thm1:bound} to conclude that:
\begin{equation} \nonumber
\int_a^b \int \Big|g_{h} (z|\bm{x}) - f_{Y(t)}(z|\bm{x}) \Big|^2 ~ d \mathbb{P}_{\bm{X}}(\bm{x})dz
\leq o_p(1) + Ch^4,
\end{equation}
\noindent where $C=\Big(\int K(u)u^2 ~du \Big)^2 \frac{2}{3}\int_a^b\int f_{Y(t)}''(z|\bm{x})^2 ~ d \mathbb{P}_{\bm{X}}(\bm{x})dz$.
\end{proof}

\subsection{Extra Clinical Trial Results} \label{sec_ex_CT}

\begin{figure}
    \centering
     \includegraphics[scale=0.5]{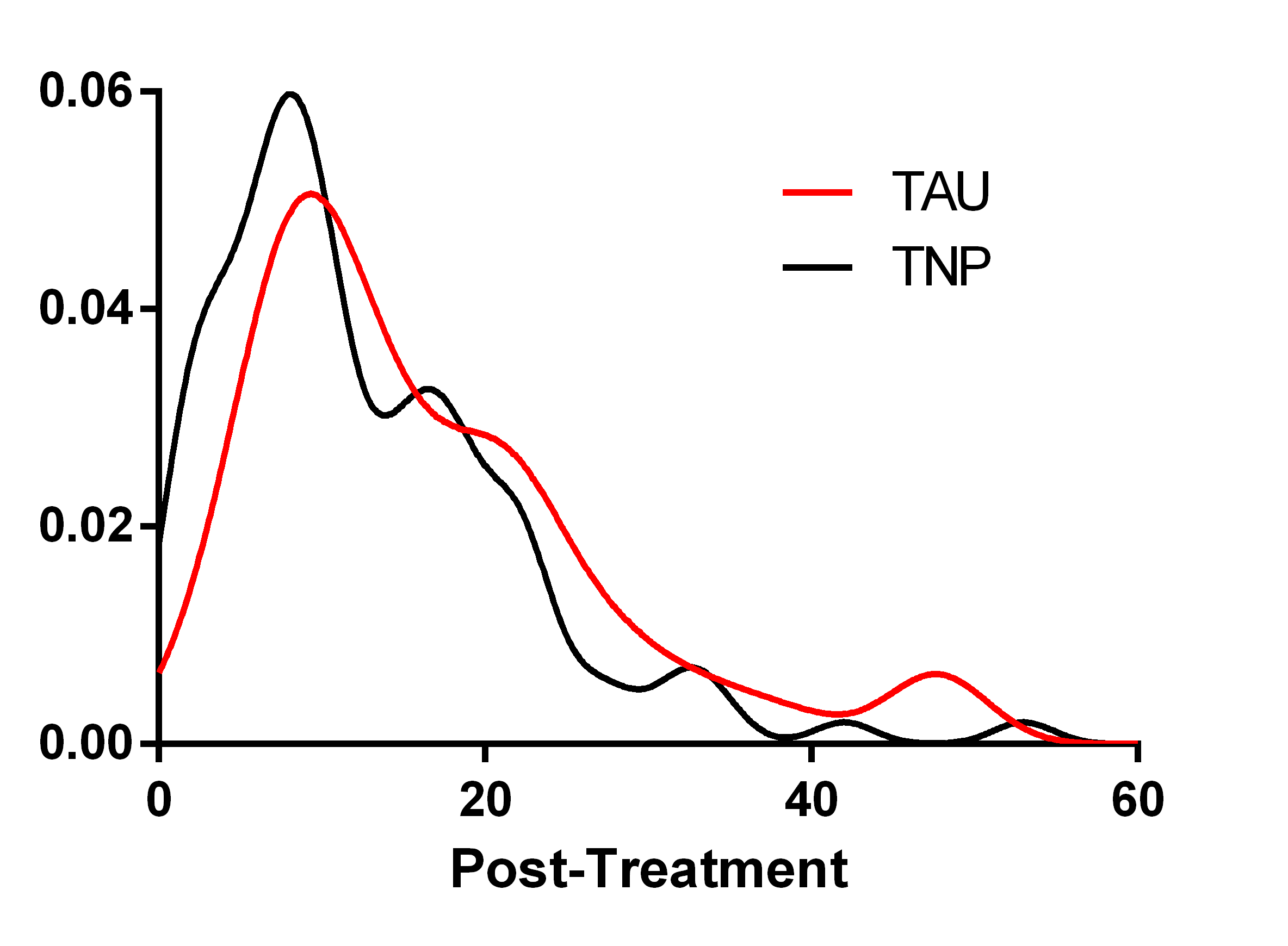}
        \caption{The unconditional densities estimated by standard kernel density estimation. The TNP density is shifted slightly to the left relative to the TAU density.} \label{fig_nicotine:all}
\end{figure}

TNPs only mildly increase smoking cessation after treatment ends. We can verify this claim by plotting the estimated unconditional densities of CO levels measured 9 weeks after completion of TNP or TAU treatment (kernel density estimation with Gaussian kernel function and unbiased cross-validation; Figure \ref{fig_nicotine:all}). Notice that the TNP density is shifted to the left relative to the TAU density, indicating that patients treated with TNP eventually smoke less than those treated with TAU on average. The difference is small at 3.97 ppm but large enough to reject the null of equality in means using a t-test (t=-2.369, p=0.020).

\subsection{Hypothesis Testing} \label{sec_hypothesis}
The TNP density in Figure 3 (b) places higher probability at lower post-treatment CO levels than the TAU density. DDR may nevertheless recover such densities often when the TNP density does \textit{not} place higher probability at lower post-treatment CO levels at the population level. We therefore also seek to reject the following null hypothesis:
\begin{equation} \nonumber
    H_0: \sup_{z} \Big[F_{Y(1)}(z|\bm{x}) - F_{Y(0)}(z|\bm{x}) \Big] \leq 0,
\end{equation}
where $F_{Y(t)}(z|\bm{X}) = \int_{-\infty}^z f_{Y(t)}(u|\bm{X})~du$. Notice that larger values of the above difference correspond to concentration of probability at lower values of $z$. We therefore reject the null when the difference is large because lower values of $z$ correspond to reduced post-treatment CO levels. We implement the hypothesis test by permuting the treatment labels, running DDR and then computing the following conditional statistic:
\begin{equation} \nonumber
    \mathcal{S} = \sup_{z}  \Big[ \widehat{F}_{h,Y(1)}(z|\bm{x}) - \widehat{F}_{h,Y(0)}(z|\bm{x}) \Big],
\end{equation}
where $\widehat{F}_{h,Y(t)}(z|\bm{X}) = \int_{-\infty}^z \widehat{g}_{h,Y(t)}(u|\bm{X})~du$. We obtained a p-value of 0.0125 with 2000 permutations for patient A by setting $T=0$ to TAU and $T=1$ to TNP. We thus reject $H_0$ in this case and conclude that the TNP density places higher probability at lower post-treatment CO levels than the TAU density. Repeating the same process with patient B on the other hand led to a p-value of 0.3635.

\end{document}